\documentclass[journal]{IEEEtran}
\usepackage{cite}
\usepackage[namelimits]{amsmath}
\usepackage{amssymb}             
\usepackage{amsfonts}            
\usepackage{mathrsfs}
\usepackage{makecell}
\usepackage{graphicx}
\usepackage{textcomp}
\usepackage{colortbl}
\usepackage{multirow, hhline}
\usepackage{multicol}
\usepackage{soul}
\usepackage{subfig}
\usepackage{float}
\usepackage[linesnumbered,ruled,vlined]{algorithm2e}
\usepackage{algpseudocode}
\usepackage[figuresleft]{rotating}

\usepackage{hyperref}
\definecolor{hl}{rgb}{0.75,0.75,0.75}
\sethlcolor{hl}

\begin{document}

\title{Towards Multi-Objective High-Dimensional Feature Selection via Evolutionary Multitasking}

\author{Yinglan~Feng,
        Liang~Feng,~\IEEEmembership{Senior Member,~IEEE,}
        Songbai~Liu,~\IEEEmembership{Member,~IEEE,}
        Sam~Kwong,~\IEEEmembership{Fellow,~IEEE,}
        and~Kay~Chen~Tan,~\IEEEmembership{Fellow,~IEEE}
        \vspace{-0.5cm}
\thanks{Y. Feng is with the Department of Computer Science, City University of Hong Kong, Hong Kong SAR (e-mail: yinglfeng2-c@my.cityu.edu.hk).}
\thanks{L. Feng is with College of Computer Science, Chongqing University, China (e-mail: liangf@cqu.edu.cn).}
\thanks{S. Liu is with College of Computer Science and Software Engineering, Shenzhen University, China (e-mail: songbai@szu.edu.cn).}
\thanks{S. Kwong is with the Department of Computer Science, City University of Hong Kong, Hong Kong SAR (e-mail: cssamk@cityu.edu.hk).}
\thanks{K.C. Tan is with the Department of Computing, The Hong Kong Polytechnic University, Hong Kong SAR (e-mail: kctan@polyu.edu.hk).}
}

\maketitle

\begin{abstract}
Evolutionary Multitasking (EMT) paradigm, an emerging research topic in evolutionary computation, has been successfully applied in solving high-dimensional feature selection (FS) problems recently. 
However, existing EMT-based FS methods suffer from several limitations, such as a single mode of multitask generation, conducting the same generic evolutionary search for all tasks, relying on implicit transfer mechanisms through sole solution encodings, and employing single-objective transformation, which result in inadequate knowledge acquisition, exploitation, and transfer.
To this end, this paper develops a novel EMT framework for multiobjective high-dimensional feature selection problems, namely MO-FSEMT. In particular, multiple auxiliary tasks are constructed by distinct formulation methods to provide diverse search spaces and information representations and then simultaneously addressed with the original task through a multi-slover-based multitask optimization scheme. Each task has an independent population with task-specific representations and is solved using separate evolutionary solvers with different biases and search preferences. A task-specific knowledge transfer mechanism is designed to leverage the advantage information of each task, enabling the discovery and effective transmission of high-quality solutions during the search process.
Comprehensive experimental results demonstrate that our MO-FSEMT framework can achieve overall superior performance compared to the state-of-the-art FS methods on 26 datasets. Moreover, the ablation studies verify the contributions of different components of the proposed MO-FSEMT.
\end{abstract}


\IEEEpeerreviewmaketitle

\section{Introduction}






Feature selection (FS) is an essential step in machine learning and data analysis that involves selecting a subset of highly informative and discriminative features (variables) from primitive data~\cite{james2013introduction}. By eliminating irrelevant or redundant features, FS contributes to improving model performance, dimensionality reduction, and enhancing computational efficiency~\cite{sammut2011encyclopedia}.
With the increasing number of features, the number of potential feature subsets grows exponentially. Due to the huge search space and complex interactions among features, FS with combinatorial characteristics remains a challenging problem, especially for high-dimensional datasets~\cite{yu2017adaptive}. 

Traditional FS approaches can be divided into four categories, including filter, wrapper, embedded, and hybrid methods combining the aforementioned techniques~\cite{tang2014feature}. 
Filter-based methods use statistical metrics, such as feature relevance or mutual information, without involving any learning algorithm to screen features~\cite{robnik2003theoretical}. Filter-based methods are usually computationally efficient and scalable to large datasets but may result in lower classification accuracy. On the other hand, wrapper-based methods evaluate feature subsets based on predictive performance by training and evaluating models on different feature subsets~\cite{kudo2000comparison,zhou2021problem,cheng2021steering}, and embedded methods integrate FS into the learning algorithm, optimizing FS and model training~\cite{wang2016sparse,liu2019embedded}, providing higher classification accuracy but computationally expensive with lower generalization performance. Hybrid methods combine FS techniques to leverage their strengths and improve overall performance~\cite{bermejo2011grasp,song2011fast,song2021fast}. 

Due to the inherent NP-hard nature of searching for the optimal subset from the feature pool, traditional methods easily fall into local optima~\cite{chandrashekar2014survey}. Therefore, in the past decade, FS methods based on evolutionary computation (EC) have gained widespread attention due to their advantages in global search capability~\cite{xue2015survey}, such as genetic algorithms (GA)~\cite{holland1992genetic}, differential evolution (DE)~\cite{das2010differential} and particle swarm optimization (PSO)~\cite{clerc2010particle}.
Regardless of GA-based~\cite{yang1998feature,khammassi2017ga}, DE-based~\cite{khushaba2011feature}, or PSO-based~\cite{tran2017new,song2020variable} FS methods, most utilize the corresponding evolutionary algorithms to search for optimal candidate feature subsets and adopt a pre-determined learning model (e.g., a classifier) to evaluate the goodness of selected features for environmental selection. Nevertheless, the extensive evaluation of solutions incurs significant computational costs, limiting their applicability in complex high-dimensional FS problems~\cite{tran2018variable}.

Furthermore, the optimization objectives of FS problems typically involve various evaluation indicators that reflect the quality of the candidate feature subsets, such as classification accuracy and the proportion of selected features. 
Most existing FS methods transform the problem into single-objective optimization by using a linear combination of different evaluation indicators~\cite{zhang2021clustering,chen2020evolutionary,chen2021evolutionary,li2023evolutionary}. To meet the requirements of diverse scenarios in real-world applications, the multiobjective FS methods offer the capability to discover the trade-off solutions between objectives without predefining weights~\cite{xue2012particle,zhou2021problem,cheng2021steering}. By producing multiple Pareto-optimal solutions in a single run, these methods provide decision-makers with diverse feature subsets, accommodating different preferences. However, these multiobjective FS methods encounter challenges when dealing with high-dimensional FS problems, suffering from slow convergence speed and early convergence to local optima.

Most recently, several FS methods based on evolutionary multitasking (EMT)~\cite{2016EMT,feng2018evolutionary,tan2021evolutionary} have been proposed to improve the effectiveness and efficiency of EC-based methods for solving high-dimensional FS problems~\cite{chen2020evolutionary,chen2021evolutionary,li2023evolutionary}.
By constructing a set of low-dimensional and related tasks within an EMT environment, these methods harness the knowledge transfer between tasks to facilitate efficient problem-solving.
For instance, in PSO-EMT~\cite{chen2020evolutionary}, an FS task with a few relevant features is created using a filtering-based approach to assist the original task. Further, multiple FS tasks are constructed based on different filter probabilities in MTPSO~\cite{chen2021evolutionary}. Subsequently, MFCSO~\cite{li2023evolutionary} explores various filter-based methods to generate auxiliary tasks.
Despite the superior performance of the EMT-based FS methods in terms of solution quality and convergence speed, their single manner of multitask generation and a generic evolutionary mechanism for all tasks limit the full acquisition and exploitation of knowledge to be transferred among tasks. Specifically, auxiliary tasks are constructed using one or more filtering-based approaches, resulting in similar search landscapes for each task and providing limited types of knowledge. Additionally, these tasks are solved by a single generic evolutionary optimizer with a population based on a unified representation, overlooking the characteristics of tasks and being constrained by the limited search capability of a single solver. Furthermore, under the existing implicit knowledge transfer mechanisms~\cite{2016EMT}, only shallow information (i.e., the encoding of solutions) can be transferred, without considering more task-specific knowledge transfer. From the perspective of feature subsets evaluation, existing EMT-based FS methods also transform the problem into single-objective optimization by using a weighted sum fitness function with a strong emphasis on the classification accuracy, which leads to a relatively larger number of selected features in the final obtained solution.

To address these issues, this paper proposes an EMT-based framework for multiobjective high-dimensional FS problems called MO-FSEMT. The proposed method generates auxiliary tasks in multiple modes to provide diverse search spaces and information representations. Then, auxiliary tasks and original tasks are optimized in parallel based on different solvers with different search preferences, guiding each task to explore promising regions and ensuring the quality and diversity of transferred knowledge. During the evolutionary process, an explicit knowledge transfer mechanism is proposed to make full use of the information carried by the transferred individuals, accelerating population convergence and helping escape potential local optima. The main contributions of this study are summarized as follows:
\begin{enumerate}
    \item A multi-manner-based problem formulation strategy, incorporating filtering-based and clustering-based methods, is designed to generate auxiliary tasks. This strategy promotes the complementary advantages of each task in finding solutions or exploring fitness landscapes and provides diverse potential information for knowledge transfer among tasks, enhancing the resolution of target FS problems.
    \item A multi-solver-based multitask optimization scheme is proposed. Each task has an independent population with task-specific representations, and a separate evolutionary solver with different biases and search preferences is employed to solve the current task. By incorporating multiple search mechanisms and facilitating knowledge transfer across tasks, high-quality solutions with diversity can be discovered and transferred during the search process.
    \item A task-specific-based knowledge transfer mechanism is introduced to fully exploit the advantageous information of tasks constructed by different problem formulations. Specifically, in addition to transferring high-dimensional solutions, solutions from filtering-based tasks provide their selected features as feature masks, while those from clustering-based tasks contribute the optimized weights (saved as the task-specific representation).
    \item A comprehensive empirical study of the proposed MO-FSEMT is conducted on 27 real high-dimensional datasets with the number of features ranging from 2308 to 49151. Experimental results demonstrate that our method is overall superior to the state-of-the-art EC-based FS methods in terms of effectiveness and efficiency.
\end{enumerate}

The remainder of this paper is organized as follows. 
Section II introduces the problem definition of multiobjective FS and provides a brief review of existing EC-based FS methods. 
Section III elaborated on the framework structure and detailed components of our proposed methods. The experimental setup and the results of comparison and ablation experiments are described in Section IV. Section V provides the conclusion of this paper.
\section{Related Work}


This section starts with the problem definition of FS. Then we provide an overview of existing EC-based FS approaches, including single-objective, multiobjective, and recent successful EMT-based FS methods. 

\subsection{Multi-Objective Feature Selection}
Feature selection is a combinatorial optimization problem aimed at choosing a subset of relevant features that optimizes given performance metrics while reducing learning costs. 
Given a labeled dataset containing $N$ samples and $D$ features, the most common formulation of the FS problem is based on selecting $d$ features ($d<D$) from the original feature set to minimize a cost function $f(.)$ (e.g., classification error rate) and selected feature size $d$. 
Specifically, a multiobjective FS problem can be expressed as follows:
\begin{equation}
\begin{aligned}
\underset{}{\min} \  & F(\boldsymbol{x}) = (f(\boldsymbol{x}), |\boldsymbol{x}| )\\
\text{s.t.}\   & \boldsymbol{x} = (x_1,x_2,\cdots, x_D) \\
 & x_i \in \{0,1\},\  i= 1,2,\cdots,D \\ 
\end{aligned}
\label{eq:fs}
\end{equation}
$\boldsymbol{x}$ indicates a solution to the FS problem, where $x_i=1$ represents that the corresponding $i$-th feature should be selected; otherwise, this feature is not selected. $|\boldsymbol{x}|$ denoted the selected feature size $d$. 

The challenge of FS lies in several aspects. 
For example, high-dimensional datasets pose a challenge due to the exponential increase in the search space, which makes it computationally expensive and time-consuming to explore all possible feature combinations. Determining an optimal subset of features that balances classification accuracy and the number of selected features is another crucial challenge. Selecting too many features may improve accuracy, but it also increases the risk of overfitting and adds additional computational burden for subsequent model training.
To address these challenges, efficient and effective FS techniques should be developed to handle high-dimensional data, account for various types of dependencies and interactions, and balance between classification accuracy and size of the selected feature subset.

\subsection{Evolutionary Feature Selection Methods}
In the past few decades, EC-based FS methods have been widely applied in the field of high-dimensional classification due to their ability to find satisfactory solutions through global search strategies within a reasonable time. The pioneering work of Siedlecki and Sklansky~\cite{siedlecki1989note} confirmed the superiority of GA over traditional non-EC FS methods. Subsequently, more GA-based FS algorithms~\cite{yang1998feature,khammassi2017ga,alivckovic2017breast} have been proposed to enhance classification performance.
PSO, as a swarm-intelligence-based solver with efficient global search capabilities, has also been widely used to solve high-dimensional FS problems. In~\cite{tran2017new}, a potential PSO (PPSO) algorithm was proposed, which adopted a discrete representation strategy to reduce the solution space. Subsequently, a variable length PSO representation (VLPSO) based on feature correlation was designed in~\cite{tran2018variable} to further enhance the performance of PPSO. These algorithms mainly improve the capability of PSO in handling FS problems, but the dimensionality of the search space is not significantly reduced. HFS-C-P~\cite{song2021fast} proposed a new three-phase hybrid FS algorithm based on correlation-guided clustering and PSO to effectively address the curse of dimensionality while considering the interaction among dimensions. Additionally, Competitive Swarm Optimizer (CSO), a variant of PSO, has been employed for high-dimensional FS problems~\cite{gu2018feature} by replacing global or personal best positions with randomly selected competitors within the swarm. However, designing genetic operators directly for the high-dimensional target task still faces challenges, such as falling into local optima. Nonetheless, the potential of exploiting the distinct search biases offered by different solvers to address high-dimensional FS problems remains to be investigated.

Furthermore, in most EC-based FS methods, the optimization objectives, such as classification accuracy and the number of selected features, are often aggregated into a single objective using weights. However, appropriate weight values setting is a challenging problem, especially when prior knowledge is not available in practical applications. Therefore, an increasing number of multiobjective FS methods have been proposed to address FS problems, aiming to obtain a balance between classification performance and the number of selected features without introducing parameters, and can provide multiple solutions in a single run. Regarding EC-based multiobjective FS algorithms, PS-NSGA~\cite{zhou2021problem} designed four novel search operators based on the Pareto dominance framework of NSGA-III~\cite{deb2013evolutionary} to enhance search capabilities and accelerate the convergence of high-dimensional FS. To enhance scalability on high-dimensional datasets, SM-MOEA~\cite{cheng2021steering} developed a dimension reduction operator and an individual repair operator based on steering matrices to reduce the number of selected features recursively. Other multiobjective FS algorithms based on different EC techniques can be found in~\cite{xue2012particle,nguyen2019multiple,hu2020multiobjective,tian2019evolutionary}. The current multiobjective FS methods need further improvement in balancing the objectives of classification accuracy and the number of selected features, especially for solving high-dimensional FS problems.

The EMT paradigm~\cite{2016EMT} is a significant branch of evolutionary transfer optimization~\cite{gupta2017insights,tan2021evolutionary,xue2023solution}, which aims to simultaneously solve multiple related optimization tasks by facilitating knowledge transfer between tasks.
Constructing an EMT optimization environment to enhance the global search capability of the target task is the latest approach to address high-dimensional FS problems. Specifically, in PSO-EMT~\cite{chen2020evolutionary}, a task with only relevant features was constructed through a single filtering method and simultaneously solved with the original task using a PSO solver with implicit knowledge transfer. For improvement, MTPSO~\cite{chen2021evolutionary} divided features into the promising feature set and the remaining feature set using the same filtering method and selected features with different probabilities to generate related FS tasks. MFCSO~\cite{li2023evolutionary} designed a multitask generation strategy with multiple filtering methods and modified a variant of CSO for multitask optimization. In general, existing EMT-based FS methods focus on the design of multitask generation strategies and mainly employ a general solver for multitask evolutionary search under implicit knowledge transfer. Considering task characteristics, diverse multitask generation and optimization with different search preferences are worth further exploration.

\section{Proposed Framework}
This section elaborates on our newly proposed EMT-based FS algorithm for solving multi-objective high-dimensional classification problems.
To be specific, we first present an overview of the MO-FSEMT framework in subsection~\ref{ch5/sec:outline}. Subsequently, the key components of MO-FSEMT are described in the following three subsections, including strategies for generating relevant FS tasks, multitask optimization design using independent populations, and knowledge transfer mechanism based on task-specific advantages.

\subsection{Framework of the proposed MO-FSEMT}\label{ch5/sec:outline}

\begin{figure}[htbp]
    \centering
    \includegraphics[width=1\linewidth]{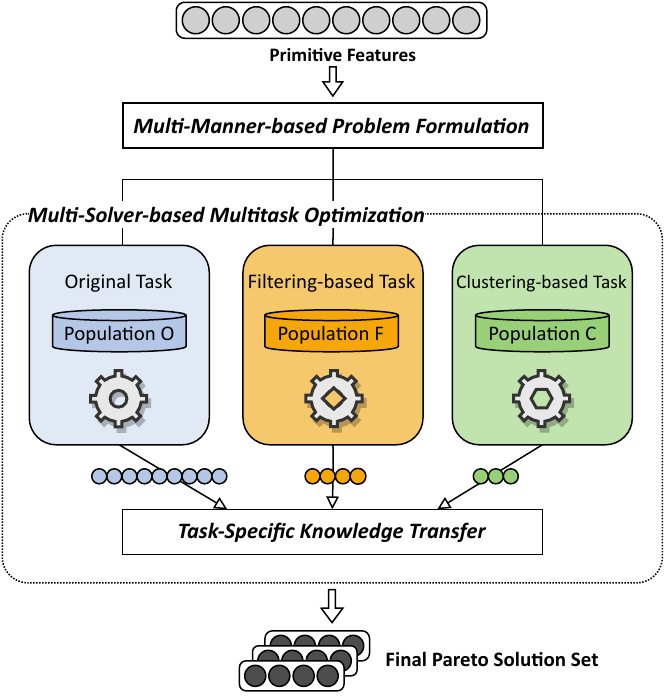}
    \caption{Illustration of the proposed MO-FSEMT framework.}
    \label{ch5/fig:framework}
\end{figure}

The overview of the proposed MO-FSEMT framework is shown in Figure~\ref{ch5/fig:framework}. The input consists of training samples, including all primitive features, while the output is the final Pareto feature sets obtained for evaluating the algorithm performance on the test set. 

The MO-FSEMT framework comprises three procedures. Firstly, the \textit{Multi-Manner-based Problem Formulation} constructs multiple distinct but highly correlated FS tasks using filtering-based and clustering-based formulation (refer to subsection~\ref{ch5/sec3/form}). Next, the original task and the generated auxiliary tasks are simultaneously addressed through \textit{Multi-Solver-based Multitask Optimization} (refer to subsection~\ref{ch5/sec3/solver}). Lastly, along with the entire evolution process, \textit{Task-Specific-based Knowledge Transfer} is conducted across tasks to enhance the overall effectiveness and efficiency of the MO-FSEMT framework (refer to subsection~\ref{ch5/sec3/transfer}). 

Algorithm~\ref{ch5/alg:framework} summarizes the pseudocode for the MO-FSEMT framework. 
Given a dataset $Data$, where $t$ represents the number of formulated auxiliary tasks, and $max_iter$ denotes the maximum number of iterations. Firstly, different tasks $T$ are generated using the proposed problem formalization strategy and are configured with task attributes to initialize their respective populations $P_T$.
Subsequently, the proposed multitask optimization scheme is utilized to update the populations and elite archives iteratively until the maximum iterations are reached. During the evolutionary process, if the elite solution remains unchanged within a given number of generations ($G$), indicating the potential entrapment in a local optimum, the proposed knowledge transfer mechanism is triggered to update the current tasks and populations. Ultimately, the elite population of all tasks is obtained as an output.

\begin{algorithm}[ht]
\caption{Framework of MO-FSEMT} 
\label{ch5/alg:framework}
\KwIn{
$Data$: the given dataset,\\
$t$: the number of formulated FS tasks,\\
$max\_iter$: maximum number of iterations.\\ 
}
\KwOut{
$Elite$: the elite population of all tasks.\\
}
$(T_O,T_F,T_C) = ProblemFormulation(Data)$\;
\For{$i=1:t$ }{
Initialize population $P_{T_i}$\;
$T_i.Elite=getElite(P_{T_i},T_i.form)$\;
}
\While{$gen\leq max\_iter$}{
$(P,T) = MultitaskOptimization(P,T)$\;
\If {transfer condition is met}{
$(P,T) = KnowledgeTransfer(P,T)$(Refer to Alg.~\ref{ch5/alg:transfer})\;
} 
}
$Elite= getElite(P)$\;
\end{algorithm}

\subsection{Problem Formulation Strategy}\label{ch5/sec3/form} 
Problem formulation is a crucial component in the framework since the effectiveness of the auxiliary tasks significantly impacts the performance of the target task.
To reduce noise and improve classification performance, we first eliminate features that are irrelevant or weakly correlated with the current classification problem to obtain the search space of the original task.
After removing irrelevant features, we conduct problem formulation in multiple manners, including filtering-based and clustering-based methods, instead of solely relying on filtering-based methods for multitask generation~\cite{chen2020evolutionary,chen2021evolutionary,li2023evolutionary}. 
In addition to facilitating the complementary advantages of different tasks in terms of optimal solutions or fitness landscapes, the proposed strategy provides diverse potential information to enhance the effectiveness of knowledge transfer among tasks, thereby promoting efficient problem-solving.
The detailed formulation process is elaborated in the following two subsections.

\subsubsection{Removing Irrelevant Features}
Due to the non-parametric nature and ability to capture nonlinear relationships among random variables~\cite{song2011fast,song2020variable}, the symmetric uncertainty (SU) is utilized in this study to measure the correlation between features and class labels, as well as among features.


For the primitive feature set $F = {f_1, f_2, ..., f_D}$, the correlation between each feature and the class label is calculated by $SU(f_i,C)$. Features with correlation values below a certain threshold are considered noise and are directly removed without further participation in subsequent feature selection.
To prevent removing relevant features by a high threshold, as well as to ensure the effectiveness of redundancy removal by avoiding setting a low value, the threshold $\rho_0$ is set as follows:
\begin{equation}
\centering
    \rho _0 = \min (\lambda \cdot SU_{\max },SU_{\lfloor D/\log D \rfloor -\textrm {th}}),
\end{equation}
where $SU_{\max }$ and $SU_{\lfloor D/\log D \rfloor -\textrm {th}}$ denote the maximum $SU(f,C)$ value of all features and the value of the $\lfloor D/\log D \rfloor$-th ranked feature, respectively. $\lambda$ is set to 0.2 in this study. 
After irrelevance reduction, the search space of the original task is based on the remaining features.

\subsubsection{Filtering-based Formulation}
By measuring the importance of features, filtering-based methods determine a feature mask, which is a binary vector indicating whether to retain features of each dimension, to reserve representative features for good classification performance. In MO-FSEMT, the filtering-based formulation utilizes the obtained feature mask to determine the simplified search space.
Different filtering methods yield distinct rankings of feature importance due to the utilization of diverse metrics for assessing the correlation between features and classes. In order to diversify the generated tasks and improve robustness, we adopt multiple filtering methods (i.e., the ReliefF algorithm~\cite{robnik2003theoretical} and feature ranking based on chi-square test~\cite{mchugh2013chi}) to evaluate feature importance. Without loss of generality, other existing filtering methods can also be involved in the feature importance ranking.

The ReliefF algorithm assigns different weights to features based on their ability to distinguish close neighbors, while feature ranking based on chi-square tests examines whether each feature is independent of the category label. The larger the associated value, the more important the feature is.  
After determining the importance of each feature based on the above filtering methods, we construct feature masks based on the knee point~\cite{zhang2014knee} scheme. Firstly, the features are sorted in descending order according to the associated values. Then, an extreme line is generated by connecting the points with maximum and minimum values. Finally, the knee point with the largest distance to the extreme line is regarded as the demarcation point to divide the feature set automatically. 

By leveraging the obtained feature mask, essential features are chosen to derive a low-dimensional search space for the task based on the filtering-based formulation, incorporating the task-specific representation.

\subsubsection{Clustering-based Formulation}
In the proposed strategy, the clustering-based formulation generates weight vectors by clustering relevant features and assigning weights to each cluster, enabling the control of a low-dimensional weight vector to accelerate the update of high-dimensional feature vectors. This work adopts the FCFC algorithm~\cite{song2021fast} to cluster relevant features and utilizes the widely-used weighted optimization (WO) method~\cite{zille2019large,feng2021multivariation} for the bidirectional mapping between the weight vector and the feature vector. Without loss of generality, other clustering mechanisms and mapping methods can also be applied to generate clustering-based tasks.

The FCFC algorithm groups similar features into the same cluster by comparing the similarity between features and known cluster centers, without predefining the number of clusters.
Details of the clustering process can be referred to~\cite{song2021fast}.

For each resulting cluster, a weight variable is assigned to control the entire set of features. 
In this study, we employ the WO method to establish a bidirectional mapping between the weight vector $\boldsymbol{u}^d$ and high-dimensional feature vectors $\boldsymbol{v}^D$, i.e., $WO(\boldsymbol{v}^D,Prime)$ and $WO\_inv(\boldsymbol{u}^d,Prime)$, to implement the simplification and reconstruction. The reference solution $Prime$ of a clustering-based task is selected from all elite archives based on the classification accuracy objective value. For more details of WO mapping, the readers are referred to the study in~\cite{feng2021multivariation}.

\subsubsection{Multitask Generation}
\begin{figure}[bp]
    \centering
    \includegraphics[width=0.95\linewidth]{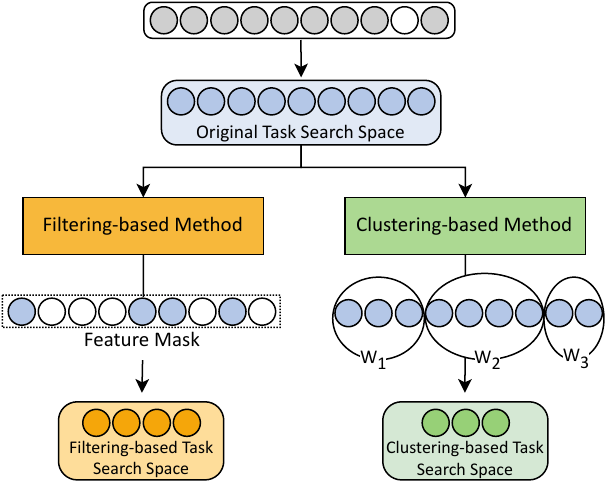}
    \caption{Toy example of the proposed problem formulation strategy.}
    \label{ch5/fig:prob_form}
\end{figure}

To understand the entire problem formulation strategy from a more intuitive perspective, a toy example is illustrated in Figure~\ref{ch5/fig:prob_form}, which showcases the process of deriving low-dimensional search spaces by two different formulations. Specifically, the task-specific representations of the filtering-based task are determined by the obtained feature mask, depicted on the left of the figure. The right side of the figure demonstrates the weight representation obtained by the clustering-based method.
Recall that in this study, the formulations are implemented using specific approaches to illustrate the effectiveness of the proposed problem formulation strategy, which can be extended to most existing filtering methods and clustering methods. 
These generated auxiliary tasks and the original task constitute a multitask optimization environment within the MO-FSEMT framework. The basic properties of each task are summarized as follows.
\begin{itemize}
    \item $Task.N$: number of the individuals belonging to the task;
    \item $Task.D$: dimension of the task-specific representation;
    \item $Task.form$: adopted problem formulation;
    \item $Task.Pop$: independent population of the task; 
    \item $Task.Elite$: elite archive of the task;
    \item $Task.mask$: feature mask obtained by filtering-based method;
    \item $Task.cluster$: clusters obtained by clustering-based method;
    \item $Task.prime$: reference solution for the clustering-based formulation;    
\end{itemize}


In MO-FSEMT, the formulation process based on each paradigm and subsequent optimization is executed independently in parallel, which allows each task to fully exploit its advantages. For example, filtering-based tasks can quickly reduce the dimension of the search space and provide a filtering scheme for high-dimensional vectors, whereas clustering-based tasks can speed up the update of decision variables and provide weight values to be applied to full-dimensional vectors.

\subsection{Multi-Solver-based Multitask Optimization}\label{ch5/sec3/solver}
In the multitask environment constructed by MO-FSEMT, in addition to the original task, there are auxiliary tasks derived from different problem formulations. Different optimization problems always have unique properties, which may require distinct solvers with varying search biases for effective problem resolution~\cite{iqbal2017cross}.
Different from most EMT-based FS methods that solve multiple tasks using a single generic evolutionary mechanism with a unified solution representation, this paper proposes a new multitask optimization scheme based on various solvers.
Each task has an independent population with task-specific representations, and a separate evolutionary solver with the specific task bias and search preference is utilized to solve the current task, thereby providing enhanced search performance. 

In our implementation, we distinguish the search preferences and problem biases for different tasks by modifying the fitness calculation approaches of CSO~\cite{cheng2014competitive}, which is a widely used search operator with a powerful exploration capability and applies pairwise competition based on fitness values to update the population. Specifically, in the independent solver of the original task, the fitness values take into account both the classification accuracy and the number of selected features. In contrast, for the auxiliary tasks, where the dimensionality of the derived search space has been significantly reduced, we only consider the objectives related to classification accuracy, enabling more effective searching on the auxiliary tasks. With distinct evolutionary search mechanisms, we can conduct task-specific searches based on the nature of each task, harnessing the advantages of different landscapes.

Minimizing the classification error rate and the number of selected features are the goals of FS problems. In this work, we use the balanced classification error rate~\cite{patterson2007fitness} and the proportion of selected features as the primary evolutionary objectives to reflect the classification performance, which are calculated by Equation~\ref{ch5/eq:error} and~\ref{ch5/eq:fr} respectively. 
\begin{equation} 
\label{ch5/eq:error}
\centering
 ErrorRate = 1-\frac {1}{C} \cdot\sum_{i=1}^{C}{TPR_{i}},
\end{equation}
where $C$ denotes the number of classes in the data samples, and $TPR_{i}$ (also known as recall) represents the proportion of correctly retrieved samples among relevant samples for class $i$. The weight of all classes is set to $\frac{1}{C}$ as the balanced accuracy has no bias among classes.
\begin{equation} 
\label{ch5/eq:fr}
\centering
 FeatureRate = \frac {|S|}{|A|},
\end{equation}
where $S$ and $A$ indicate the number of selected features and full features, respectively.
Besides, in the context of multiobjective fitness calculation for auxiliary tasks, an additional objective concerning classification accuracy is incorporated as an assistant metric. The additional objective is derived by replacing $TPR$ in Equation~\ref{ch5/eq:error} with precision values, quantifying the ratio of truly relevant samples among all the retrieved samples.


\linespread{1}
\begin{algorithm}[htbp]
\caption{Update Elite Archive} 
\label{ch5/alg:getElite}
\KwIn{
$Population$: the current population, $N$: the limitation of the elite archive.\\ 
}
\KwOut{
$Elite$: the obtained elite archive.\\
}
Initialize $Next$ as a zero vector representing the individual selection\;
$\textit{FrontNo1} = \textit{NDSort}([obj_{ErrorRate},obj_{FeatureRate}])$\;
$\textit{FrontNo2} = \textit{NDSort}([obj_{ErrorRate},obj_{Assistant}])$\;
Individuals on non-dominated front in $\textit{FrontNo1}$ are selected into $Next$ with a probability $p=\textit{Normalization}(obj_{ErrorRate})$\;
Individuals on non-dominated front of $\textit{FrontNo2}$ are directly selected into $Next$\;
\If {$\sum {Next} > N$) }{
Sort \textit{Population(Next)} based on $obj_{ErrorRate}$, and save top-$N$ into $Elite$\;
}
\Else{
$Elite = Population(Next)$\;
}
\end{algorithm}

In addition to fitness calculation, the preference for classification accuracy is also reflected in the selection of the elite population, as shown in Algorithm~\ref{ch5/alg:getElite}. Without any preference, an increasing number of individuals with few features but high error rates would dominate the majority of the population, leading to a decrease in average classification accuracy and even falling into local optima. Therefore, following~\cite{yang2020cars}, which merges different Pareto stages to keep solutions with a preference for a certain objective, we perform two rounds of non-dominated sorting for both the primary objectives and accuracy-related objectives (from lines 2 to 3).
For the former with feature rates, solutions are selected from the non-dominated front with a certain probability $p$, which is the normalized objective value of the error rate. As for the accuracy-related objectives, we retain all the solutions from the non-dominated front (from lines 4 to 5). This approach ensures that individuals with higher accuracy can be preserved in the population, reducing the possibility of being trapped in local optima. When the number of selected solutions exceeds the archive capacity, individuals with lower accuracy are eliminated (line 7). For a detailed description of the elite population selection process, please refer to Algorithm~\ref{ch5/alg:getElite}.


\subsection{Task-Specific-based Knowledge Transfer}\label{ch5/sec3/transfer}
Effective knowledge transfer mechanisms can enhance the overall search capability of the algorithm, accelerate convergence, and reduce computation time, which is one of the core components of the MO-FSEMT framework. 
In order to fully leverage the advantages of tasks, we propose a task-specific-based knowledge transfer mechanism, as shown in the Algorithm~\ref{ch5/alg:transfer}. 

\begin{algorithm}[ht]
\caption{Task-Specific Knowledge Transfer} 
\label{ch5/alg:transfer}
\KwIn{All tasks with their properties of $Task.Pop$, $Task.Elite$, $Task.form$, and $Task.N$; and the random transfer probability $rtp$.\\}
\KwOut{Updated Tasks.\\}
\For{$T_i$ \textbf{in} $Tasks$}
{
\rm Integrating the elite archive of other tasks as $candiPop$ to be transferred\;
\For{$n = 1:T_i.N$}
{
\rm Select the $n$-th individual from $T_i.Pop$ as $P_1$ with the high-dimensional
solution $\boldsymbol{v}^D$ and the task-specific representation $\boldsymbol{u}^d$\;
\If{$rand(0,1)<rtp$}
{
\rm Randomly find a individual from $candiPop$ as $P_2$\;
\If {$P_2 \in Task.form = F$ (Filtering to Others)}
{
Apply $logical(P_2.\boldsymbol{v})$ (as the feature mask) to $P_1.\boldsymbol{v}$ in the original search space\;
$P_{new}(n).\boldsymbol{v} = logical(P_2.\boldsymbol{v}) \land P_1.\boldsymbol{v}$\;
}
\ElseIf{$P_2 \in Task.form = C$ (Clustering to Others)}
{
Apply $P_2.\boldsymbol{u}$ (as the weight vector) to $P_1.\boldsymbol{v}$ (as the reference solution) in the original search space\;
$P_{new}(n).\boldsymbol{v} = WO\_inv(P_2.\boldsymbol{u},P_1.\boldsymbol{v})$\;
}
\Else{
$P_2\in Task.form = O$ \textit{(Original to Others)} provides decision variables with full features and is used as a reference solution to apply the above process in the search space of $P_1$\;
}
}
$P_{new}(n) = Mutate(P_1)$\;
}
$T_i.Pop = \textit{EnvironmentalSelection}(T_i.Pop \cup P_{new})$\;
$T_i.Elite = \textit{getElite}(T_i.Elite \cup T_i.Pop)$.
}
\end{algorithm}

Given all tasks with their latest properties of $Task.Pop$, $Task.Elite$, $Task.form$, and so on, knowledge transfer is performed across all tasks in turn.
When transferring knowledge to the current task, elite archives of other tasks are integrated as the candidate population for the transfer (line 2).
During the process of updating individual, including the high-dimensional solution $\boldsymbol{v}^D$ for evaluation and the task-specific representation $\boldsymbol{u}^d$, a random transfer probability ($rtp$) is predefined to control whether to transfer knowledge from other tasks. A larger $rtp$ encourages more transmission from other tasks. For simplicity, the value of $rtp$ is set to 0.6, according to the suggestion in~\cite{chen2021evolutionary}. We generate a random number between 0 and 1. If it is less than the predefined $rtp$ (the proposed knowledge transfer strategy is activated), an individual is selected randomly from the candidate population for the follow-up knowledge transfer (line 6).

Continuous real-valued encoding~\cite{chen2021evolutionary} is commonly adopted in FS tasks instead of binary representation. The range of each dimension is set between 0 and 1, with a threshold value $\theta$ determining the selection of features.
Under the real-valued encoding, when knowledge is transferred from a filtering-based task to other tasks (from lines 8 to 9), the transferred individual $P_2$ not only provides the encoding of the high-dimensional solution but also offers important information through $\theta$-determined feature selection (indicated by a logical function). In this case, the feature selection of $P_2$ is regarded as a feature mask to filter the high-dimensional solution of the current individual $P_1$, obtaining a new solution. When knowledge transfer occurs between a clustering-based task and other tasks (from lines 11 to 12), the transferred individual $P_2$ primarily contributes weight information (stored in the task-specific representation $\boldsymbol{u}^d$). In this scenario, the new solution is generated by the inverse mapping of WO, where weights of $P_2$ are applied to the high-dimensional solution of the current individual $P_1$, which serves as the reference solution. 

\begin{figure}[htbp]
    \centering
    \includegraphics[width=0.95\linewidth]{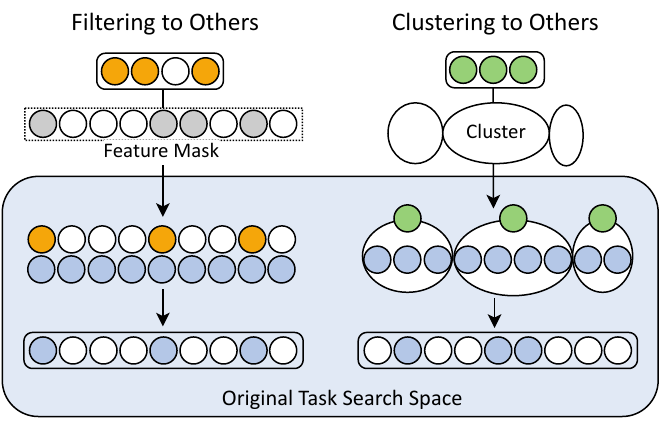}
    \caption{Toy example of the proposed task-specific based knowledge transfer mechanism.}
    \label{ch5/fig:know_trans}
\end{figure}

Figure~\ref{ch5/fig:know_trans} gives a toy example to depict the representation of different types of information and their interaction more intuitively. It is worth noting that both of the above cases involve knowledge transfer in the high-dimensional space of the original task. On the contrary, when the transferred individual $P_2$ originates from the original task (line 14), the knowledge transfer occurs in the auxiliary task space where the current individual $P_1$ resides. In this situation, the decision variables with full features carried by the individual $P_2$ are most valuable for other tasks. Therefore, the high-dimensional solution of $P_2$ is used as the reference solution and undergoes the aforementioned process according to the problem formulation of $P_1$. After completing the knowledge transfer for $P_1$, polynomial mutation (PM)~\cite{deb2014analysing} is applied to evolve the new individual in line 15. Finally, the population and the elite archive of each task are updated sequentially (from lines 16 to 17).

Instead of relying solely on the high-dimensional encoding of solutions for information sharing through crossover operators, the proposed knowledge transfer mechanism fully exploits task-specific information, such as the selection of the further reduced feature sets of individuals on filter-based tasks and the weight information of individuals on clustering-based tasks.
By effectively harnessing the strengths and potential information from different tasks, this approach enables efficient knowledge transfer and improves the quality of solutions.

\section{Experimental Study}
In this section, we present the experimental study conducted to investigate the effectiveness and efficiency of the proposed MO-FSEMT framework in addressing the high-dimensional FS problems. Specifically, we first describe the experimental setup, including the real-world datasets, comparison methods, and the adopted parameter settings. Then, a detailed comparison and analysis of the experimental results are provided. Finally, the effectiveness of the strategies designed in the MO-FSEMT framework is validated through ablation experiments.

\subsection{Datasets}
For the experimental evaluation, we utilize 26 high-dimensional datasets with the number of features ranging from 2308 to 22283 and a higher-dimensional case involving 49151 features. These datasets are available at \url{http://featureselection.asu.edu} and \url{https://github.com/gelin123}. Table~\ref{ch5/tab:dataset} summarizes the basic information of the adopted datasets, including the number of features, instances, and categories, as well as the proportions of the smallest and largest categories. It is worth noting that the distribution of most datasets is highly imbalanced, posing challenges for accurate feature selection. Furthermore, several datasets with balanced category distribution are also involved in the comparison to assess the classification performance of MO-FSEMT when handling diverse datasets.

\linespread{0.9}
\begin{table}[tbp]
  \centering
  \caption{Properties of the adopted datasets}
  \scalebox{0.85}{
    \begin{tabular}{lrrrrr}
    \hline
    \hline
    \multicolumn{1}{l}{\multirow{2}[2]{*}{\textbf{Datasets}}} & \multicolumn{1}{c}{\multirow{2}[2]{*}{\textbf{\#Feature}}} & \multicolumn{1}{c}{\multirow{2}[2]{*}{\textbf{\#Instance}}} & \multicolumn{1}{c}{\multirow{2}[2]{*}{\textbf{\#Category}}} & \multicolumn{1}{c}{\textbf{\%Smallest}} & \multicolumn{1}{c}{\textbf{\% Largest}} \\
          &  -   &  -   &  -   & \multicolumn{1}{c}{\textbf{category}} & \multicolumn{1}{c}{\textbf{category}} \\
    \hline
    11\_Tumors & 12533 & 174   & 11    & 3.4   & 15.5 \\
    9\_Tumors & 5726  & 60    & 9     & 3.3   & 15.0 \\
    Adenocarcinoma & 9672  & 76    & 2     & 15.8  & 84.2 \\
    ALLAML & 7129  & 72    & 2     & 34.7  & 65.3 \\
    AR10P & 2400  & 130   & 10    & 10.0  & 10.0 \\
    Brain\_Tumor1 & 5920  & 90    & 5     & 4.4   & 66.7 \\
    Brain\_Tumor2 & 10367 & 50    & 4     & 14.0  & 30.0 \\
    Breast & 4869  & 95    & 3     & 18.9  & 46.3 \\
    CLL-SUB-111 & 11340 & 111   & 3     & 9.9   & 45.9 \\
    DLBCL & 5469  & 77    & 2     & 24.7  & 75.3 \\
    GLI-85  & 22283 & 85    & 2     & 30.6  & 69.4 \\
    Leukemia1 & 5327  & 72    & 3     & 12.5  & 52.8 \\
    Leukemia2 & 11225 & 72    & 3     & 27.8  & 38.9 \\
    Lung\_Cancer & 12600 & 203   & 5     & 3.0   & 68.5 \\
    Lymphoma & 4026  & 62    & 3     & 14.5  & 67.7 \\
    Nci   & 5244  & 61    & 8     & 8.2   & 14.8 \\
    Nci9  & 9712  & 60    & 9     & 3.3   & 15.0 \\
    ORL10P  & 10304 & 100   & 10    & 10.0  & 10.0 \\
    PIE10P  & 2420  & 210   & 10    & 10.0  & 10.0 \\
    PIX10P  & 10000 & 100   & 10    & 10.0  & 10.0 \\
    Prostate\_GE  & 5966  & 102   & 2     & 49.0  & 51.0 \\
    Prostate\_Tumor & 10509 & 102   & 2     & 49.0  & 51.0 \\
    Prostate\_6033 & 6033  & 102   & 2     & 49.0  & 51.0 \\
    SMK-CAN-187  & 19993 & 187   & 2     & 48.1  & 51.9 \\
    SRBCT & 2308  & 83    & 4     & 13.3  & 34.9 \\
    TOX-171  & 5748  & 171   & 4     & 22.8  & 26.3 \\
    Case\_Study  & 49151  & 180   & 4     & 12.8  & 45.0 \\
    \hline
    \hline
    \end{tabular}%
    }
  \label{ch5/tab:dataset}%
\end{table}%

\subsection{Compared Methods and Parameter Settings}

\linespread{0.9}
\begin{table*}[tbp]
  \centering
  \caption{The statistics of the average classification results obtained by seven FS methods on the 26 datasets. The best performance is marked by boldface.}
    \scalebox{0.9}{
    \begin{tabular}{p{2.6cm}rrrrrrr}
    \hline
    \hline
    \textbf{Dataset} & \multicolumn{1}{c}{\textbf{PS-NSGA}} & \multicolumn{1}{c}{\textbf{SM-MOEA}} & \multicolumn{1}{c}{\textbf{HFS-C-P}} & \multicolumn{1}{c}{\textbf{PSO-EMT}} & \multicolumn{1}{c}{\textbf{MTPSO}} & \multicolumn{1}{c}{\textbf{MF-CSO}} & \multicolumn{1}{c}{\textbf{MO-FSEMT}} \\
    \hline
    11\_Tumors & 81.53 (2.69) - & 67.16 (16.53) - & 79.18 (1.23) - & 82.16 (2.06) - & 81.86 (0.98) - & 79.25 (1.86) - & \textbf{85.18 (1.01)} \\
    9\_Tumors & 49.30 (4.79) - & 24.51 (14.87) - & 48.68 (5.04) - & 54.42 (3.40) - & 56.76 (2.70) - & 47.18 (4.31) - & \textbf{68.27 (1.98)} \\
    Adenocarcinoma & 59.87 (3.89) - & 53.06 (9.22) - & 63.48 (5.78) - & \textbf{70.25 (4.67) =} & 69.24 (5.40) = & 60.08 (3.81) - & 68.68 (2.98) \\
    ALLAML & 88.32 (4.05) - & 75.23 (15.98) - & 83.58 (4.63) - & 89.91 (2.23) - & 93.37 (1.42) - & 92.26 (2.95) - & \textbf{96.45 (0.51)} \\
    AR10P & \textbf{88.15 (2.72) +} & 49.66 (13.59) - & 73.15 (2.87) - & 72.67 (2.28) - & 72.73 (1.87) - & 83.42 (2.08) + & 81.21 (1.12) \\
    Brain\_Tumor1 & 62.78 (3.96) - & 57.43 (14.67) - & 62.20 (3.55) - & \textbf{76.44 (3.96) +} & 76.17 (2.49) = & 68.68 (3.09) - & 74.70 (2.24) \\
    Brain\_Tumor2 & 66.98 (5.43) - & 58.62 (18.62) - & 71.70 (4.47) - & 71.82 (3.31) - & 72.87 (2.84) - & 71.89 (4.97) - & \textbf{82.28 (1.93)} \\
    Breast3 & 55.57 (4.32) - & 44.67 (13.81) - & 54.58 (3.62) - & 57.07 (3.63) - & 56.20 (2.27) - & 52.60 (4.73) - & \textbf{63.85 (2.58)} \\
    CLL\_SUB\_111 & 74.16 (3.82) - & 60.08 (18.10) - & 69.15 (3.60) - & 66.32 (2.77) - & 66.16 (2.99) - & 76.60 (2.05) - & \textbf{80.51 (1.48)} \\
    DLBCL & 83.37 (5.11) - & 69.85 (12.34) - & 83.51 (4.80) - & 91.50 (1.78) - & 91.28 (1.26) - & 92.70 (3.26) - & \textbf{95.48 (0.57)} \\
    GLI\_85 & 80.67 (4.48) - & 80.59 (17.37) - & 77.41 (4.25) - & 83.38 (3.89) - & 85.09 (2.93) - & 83.88 (2.84) - & \textbf{93.33 (1.51)} \\
    Leukemia1 & 86.10 (4.10) - & 81.12 (15.83) - & 88.57 (3.90) - & 90.76 (1.74) - & 92.25 (1.01) - & 89.60 (2.91) - & \textbf{95.35 (0.42)} \\
    Leukemia2 & 86.78 (4.33) - & 68.94 (16.76) - & 89.57 (3.59) - & 94.47 (2.06) - & 93.24 (1.52) - & 91.66 (2.58) - & \textbf{96.11 (0.58)} \\
    Lung\_Cancer & 84.88 (3.03) - & 78.08 (12.51) - & 79.44 (4.13) - & 86.19 (2.00) - & 88.61 (1.57) - & 83.30 (3.85) - & \textbf{90.77 (1.14)} \\
    Lymphoma & 93.71 (4.57) - & 90.14 (13.13) = & 94.48 (3.49) - & 98.95 (0.43) = & 98.94 (0.53) = & \textbf{99.02 (1.42) +} & 98.75 (0.44) \\
    Nci   & 64.56 (5.22) - & 31.81 (17.74) - & 67.29 (4.94) - & \textbf{74.79 (2.98) =} & 73.86 (2.18) = & 68.05 (3.47) - & 74.01 (1.62) \\
    Nci9  & 51.86 (4.23) - & 19.35 (15.90) - & 42.00 (3.22) - & 63.11 (3.44) - & 63.58 (2.50) - & 44.92 (2.79) - & \textbf{76.64 (2.40)} \\
    ORL10P & 90.35 (2.87) - & 74.29 (15.13) - & 95.20 (1.47) - & 96.94 (0.53) - & 95.60 (1.02) - & 95.75 (1.48) - & \textbf{99.30 (0.20)} \\
    PIE10P & 95.26 (1.08) - & 62.38 (15.10) - & 97.86 (0.84) - & 98.65 (0.51) = & \textbf{99.18 (0.52) +} & 99.05 (0.67) + & 98.58 (0.25) \\
    PIX10P & 94.05 (1.54) - & 87.32 (8.29) - & 98.00 (1.38) = & 98.79 (0.47) = & \textbf{98.98 (0.38) =} & 98.00 (0.79) - & 98.95 (0.15) \\
    Prostate\_GE & 87.01 (2.63) - & 82.95 (7.33) - & 79.24 (3.40) - & 86.51 (1.40) - & 87.78 (1.36) - & 83.02 (1.70) - & \textbf{92.12 (0.71)} \\
    Prostate\_Tumor & 88.88 (2.99) - & 83.13 (9.64) - & 78.11 (3.02) - & 81.50 (2.07) - & 84.90 (1.48) - & 89.34 (2.39) - & \textbf{91.34 (1.20)} \\
    Prostate\_6033 & 87.22 (3.38) - & 85.36 (9.42) = & 87.54 (1.52) - & 86.76 (1.53) - & 87.26 (0.99) - & 87.67 (2.52) - & \textbf{91.80 (0.85)} \\
    SMK\_CAN\_187 & 61.74 (3.25) - & 61.09 (9.51) - & 63.52 (2.07) - & 65.17 (1.55) - & 64.25 (1.64) - & 65.00 (1.77) - & \textbf{73.59 (1.74)} \\
    SRBCT & 91.00 (2.25) - & 68.07 (15.56) - & 97.75 (1.79) = & 96.49 (1.21) - & 97.18 (0.60) - & 96.50 (1.95) - & \textbf{98.58 (0.21)} \\
    TOX\_171 & 93.71 (1.89) + & 72.16 (13.37) - & 75.82 (2.65) - & 83.36 (2.14) - & 83.56 (1.89) - & \textbf{94.55 (0.88) +} & 88.60 (0.98)\\
    \hline
    \multicolumn{1}{c}{+/-/=} & \multicolumn{1}{c}{2/24/0} & \multicolumn{1}{c}{0/24/2} & \multicolumn{1}{c}{0/24/2} & \multicolumn{1}{c}{1/20/5} & \multicolumn{1}{c}{1/20/5} & \multicolumn{1}{c}{4/22/0} &  \\
    \hline
    \hline
    \end{tabular}%
    }
  \label{ch5/tab:meanAcc_all}%
\end{table*}%

To assess the performance of the proposed MO-FSEMT, we compared it with six competitive evolutionary FS algorithms, namely: 1) PS-NSGA~\cite{zhou2021problem}; 2) SM-MOEA~\cite{cheng2021steering}; 3) HFS-C-P~\cite{zhang2021clustering}; 4) PSO-EMT~\cite{chen2020evolutionary}; 5) MTPSO~\cite{chen2021evolutionary}; 6) MFCSO~\cite{li2023evolutionary}.
Specifically, PS-NSGA and SM-MOEA are EC-based multiobjective optimization algorithms that employ wrapper-based FS strategies. The latter four comparison algorithms convert multiple objectives into single-objective optimization through weighted sum. The difference is that HFS-C-P employs a three-phase hybrid FS algorithm based on correlation-guided filtering, clustering, and PSO wrapper. On the other hand, PSO-EMT, MTPSO, and MFCSO utilize single or multiple filtering methods to construct auxiliary tasks and combine them with PSO or CSO wrapper for multitask evolutionary optimization. These methods are chosen to represent optimization algorithms with different numbers of objectives, different categories of FS techniques, and different task construction strategies, providing a comprehensive evaluation of the MO-FSEMT framework. All the source codes of the compared algorithms are provided by the original references. Since the performance improvements upon the underlying basic FS methods or optimizers by the comparison algorithms have been empirically confirmed in the related literature, this study directly explores the advantages of MO-FSEMT over the existing state-of-the-art FS methods, omitting the comparison with the basic techniques.

For a fair comparison, the maximum number of generations for all algorithms in the experiments was fixed at $iter = 100$. Due to the varying number of features in the datasets presented in Table~\ref{ch5/tab:dataset}, the population size $N$ for all algorithms is set to 1/30 of the feature count and limited to a maximum of 200 to maintain a certain search efficiency and computational cost. 
For the chosen six comparison algorithms, we adopt the parameter settings according to the suggestions of their corresponding references. It is worth noting that, consistent with PSO-EMT, MTPSO, and MF-CSO, the proposed MO-FSEMT method uses a continuous real encoding instead of binary representation for feature selection. Each dimension in MO-FSEMT ranges between [0,1] and the selected features are determined based on a threshold value, denoted as $\theta$. Following the recommendation in~\cite{chen2021evolutionary}, the threshold value is set to $\theta=0.6$. In MO-FSEMT, the number of tasks is set to $t=5$, including the original FS task, two different filtering-based tasks, and two different clustering-based tasks. Furthermore, due to the simplicity and effectiveness of the K-nearest neighbors (KNN) algorithm, it is used as the classifier to evaluate individual fitness and verify the performance of each algorithm in the experiments, with $k=5$. To ensure the fairness and validity of the comparison results, all algorithms in the experiments adopt the ten-fold cross-validation method to verify the final results, whereas a five-fold cross-validation is applied to the training set to evaluate the classification performance of the current individual during the evolutionary process.

\begin{table*}[!ht]
  \centering
  \caption{The statistics of the best classification results obtained by seven FS methods on the 26 datasets. The best performance of each dataset is marked by boldface.}
    \scalebox{0.9}{
    \begin{tabular}{p{2.6cm}rrrrrrr}
    \hline
    \hline
    \textbf{Dataset} & \multicolumn{1}{c}{\textbf{PS-NSGA}} & \multicolumn{1}{c}{\textbf{SM-MOEA}} & \multicolumn{1}{c}{\textbf{HFS-C-P}} & \multicolumn{1}{c}{\textbf{PSO-EMT}} & \multicolumn{1}{c}{\textbf{MTPSO}} & \multicolumn{1}{c}{\textbf{MF-CSO}} & \multicolumn{1}{c}{\textbf{MO-FSEMT}} \\
    \hline
    11\_Tumors & 81.53 (2.69) - & 84.56 (10.14) - & 79.18 (1.23) - & 86.05 (1.99) - & 90.57 (1.10) - & 79.25 (1.86) - & \textbf{92.70 (1.48)} \\
    9\_Tumors & 49.30 (4.79) - & 58.63 (18.16) - & 48.68 (5.04) - & 61.97 (4.00) - & 70.93 (3.71) - & 47.18 (4.31) - & \textbf{82.01 (2.67)} \\
    Adenocarcinoma & 59.87 (3.89) - & 70.00 (13.89) - & 63.48 (5.78) - & 75.70 (4.76) - & 76.99 (5.45) - & 60.08 (3.81) - & \textbf{84.62 (5.50)} \\
    ALLAML & 88.32 (4.05) - & 92.08 (12.89) - & 83.58 (4.63) - & 95.68 (1.62) - & 98.99 (1.18) - & 92.26 (2.95) - & \textbf{100.00 (0.00)} \\
    AR10P & \textbf{88.15 (2.72) =} & 83.10 (9.33) = & 73.15 (2.87) - & 83.34 (2.60) - & 80.27 (2.06) - & 83.42 (2.08) - & 87.80 (1.71) \\
    Brain\_Tumor1 & 62.78 (3.96) - & 76.50 (17.86) - & 62.20 (3.55) - & 80.50 (3.90) - & 86.36 (3.36) = & 68.68 (3.09) - & \textbf{87.33 (3.85)} \\
    Brain\_Tumor2 & 66.98 (5.43) - & 86.67 (20.66) = & 71.70 (4.47) - & 78.28 (3.98) - & 88.19 (3.07) - & 71.89 (4.97) - & \textbf{93.48 (1.85)} \\
    Breast3 & 55.57 (4.32) - & 63.86 (11.22) - & 54.58 (3.62) - & 62.87 (3.61) - & 69.73 (2.66) - & 52.60 (4.73) - & \textbf{77.07 (3.00)} \\
    CLL\_SUB\_111 & 74.16 (3.82) - & 82.64 (9.70) - & 69.15 (3.60) - & 71.51 (3.54) - & 76.58 (2.94) - & 76.60 (2.05) - & \textbf{89.70 (1.79)} \\
    DLBCL & 83.37 (5.11) - & 90.21 (9.77) - & 83.51 (4.80) - & 96.34 (1.63) - & 99.17 (0.75) - & 92.70 (3.26) - & \textbf{100.00 (0.00)} \\
    GLI\_85 & 80.67 (4.48) - & 95.30 (6.14) = & 77.41 (4.25) - & 89.38 (4.46) - & 94.88 (1.80) - & 83.88 (2.84) - & \textbf{98.71 (0.92)} \\
    Leukemia1 & 86.10 (4.10) - & \textbf{100.00 (0.00)} = & 88.57 (3.90) - & 95.04 (2.12) - & 98.07 (0.87) - & 89.60 (2.91) - & \textbf{100.00 (0.00)} \\
    Leukemia2 & 86.78 (4.33) - & 89.58 (15.08) - & 89.57 (3.59) - & 96.69 (1.55) - & 97.91 (1.46) - & 91.66 (2.58) - & \textbf{100.00 (0.00)} \\
    Lung\_Cancer & 84.88 (3.03) - & 88.94 (11.42) = & 79.44 (4.13) - & 89.84 (1.72) - & 95.97 (1.56) - & 83.30 (3.85) - & \textbf{97.16 (0.90)} \\
    Lymphoma & 93.71 (4.57) - & 96.67 (8.72) = & 94.48 (3.49) - & 99.07 (0.53) - & 99.78 (0.47) - & 99.02 (1.42) - & \textbf{100.00 (0.00)} \\
    Nci   & 64.56 (5.22) - & 68.50 (11.62) - & 67.29 (4.94) - & 78.89 (3.05) - & 83.96 (2.75) - & 68.05 (3.47) - & \textbf{86.62 (2.49)} \\
    Nci9  & 51.86 (4.23) - & 51.00 (15.53) - & 42.00 (3.22) - & 71.80 (3.45) - & 73.14 (3.57) - & 44.92 (2.79) - & \textbf{89.91 (2.59)} \\
    ORL10P & 90.35 (2.87) - & 98.61 (3.55) = & 95.20 (1.47) - & 99.36 (0.78) - & 97.26 (0.80) - & 95.75 (1.48) - & \textbf{100.00 (0.00)} \\
    PIE10P & 95.26 (1.08) - & 99.50 (1.31) = & 97.86 (0.84) - & 99.56 (0.52) = & 99.64 (0.44) = & 99.05 (0.67) - & \textbf{99.77 (0.24)} \\
    PIX10P & 94.05 (1.54) - & 99.48 (1.59) - & 98.00 (1.38) - & 98.84 (0.44) - & 98.98 (0.38) - & 98.00 (0.79) - & \textbf{99.55 (0.51)} \\
    Prostate\_GE & 87.01 (2.63) - & 91.75 (6.24) - & 79.24 (3.40) - & 90.66 (1.86) - & 92.15 (1.34) - & 83.02 (1.70) - & \textbf{97.27 (0.86)} \\
    Prostate\_Tumor & 88.88 (2.99) - & 94.00 (6.81) = & 78.11 (3.02) - & 86.38 (2.45) - & 92.51 (1.19) - & 89.34 (2.39) - & \textbf{96.57 (1.59)} \\
    Prostate\_6033 & 87.22 (3.38) - & 95.37 (5.68) = & 87.54 (1.52) - & 90.53 (1.43) - & 91.91 (0.88) - & 87.67 (2.52) - & \textbf{97.04 (0.94)} \\
    SMK\_CAN\_187 & 61.74 (3.25) - & 74.90 (4.73) - & 63.52 (2.07) - & 69.43 (1.74) - & 72.72 (1.91) - & 65.00 (1.77) - & \textbf{81.90 (2.05)} \\
    SRBCT & 91.00 (2.25) - & 96.67 (6.14) - & 97.75 (1.79) - & 98.91 (0.85) - & 99.94 (0.25) = & 96.50 (1.95) - & \textbf{100.00 (0.00)} \\
    TOX\_171 & 93.71 (1.89) = & 92.08 (7.01) = & 75.82 (2.65) - & 89.07 (2.07) - & 93.45 (1.74) - & \textbf{94.55 (0.88) =} & 94.46 (1.04) \\
    \hline
    \multicolumn{1}{c}{+/-/=}& \multicolumn{1}{c}{0/24/2} & \multicolumn{1}{c}{0/15/11} & \multicolumn{1}{c}{0/26/0} & \multicolumn{1}{c}{0/25/1} & \multicolumn{1}{c}{0/23/3} & \multicolumn{1}{c}{0/25/1} &  \\
    \hline
    \hline
    \end{tabular}%
    }
  \label{ch5/tab:maxAcc_all}%
\end{table*}%

\begin{table*}[ht]
  \centering
  \caption{The statistics of the number of the features selected by seven FS methods on the 26 datasets. The best performance of each dataset is marked by boldface.}
  \scalebox{0.95}{
    \begin{tabular}{lrrrrrrr}
    \hline
    \hline
    \textbf{Dataset} & \multicolumn{1}{c}{\textbf{PS-NSGA}} & \multicolumn{1}{c}{\textbf{SM-MOEA}} & \multicolumn{1}{c}{\textbf{HFS-C-P}} & \multicolumn{1}{c}{\textbf{PSO-EMT}} & \multicolumn{1}{c}{\textbf{MTPSO}} & \multicolumn{1}{c}{\textbf{MF-CSO}} & \multicolumn{1}{c}{\textbf{MO-FSEMT}} \\
    \hline
    11\_Tumors & 159.05 + & \textbf{13.54 +} & 10286.65 - & 508.19 - & 1444.92 - & 623.55 - & 271.78\\
    9\_Tumors & 121.35 = & \textbf{4.03 +} & 4598.65 - & 243.92 - & 450.34 - & 130.80 = & 113.48 \\
    Adenocarcinoma & 15.95 = & \textbf{3.23 +} & 634.10 - & 353.53 - & 254.60 - & 231.00 - & 21.59 \\
    ALLAML & 2.90 + & \textbf{2.64 +} & 132.15 - & 210.15 - & 371.85 - & 80.65 - & 60.61 \\
    AR10P & 21.10 + & \textbf{2.93 +} & 164.60 - & 225.90 - & 283.81 - & 121.00 - & 60.90 \\
    Brain\_Tumor1 & 33.70 + & \textbf{4.13 +} & 925.20 - & 340.25 - & 639.09 - & 118.50 - & 50.17 \\
    Brain\_Tumor2 & 13.20 + & \textbf{3.49 +} & 1614.50 - & 505.65 - & 1152.68 - & 159.40 = & 111.61 \\
    Breast3 & 95.15 - & \textbf{3.44 +} & 324.15 - & 263.82 - & 465.12 - & 154.50 - & 46.64 \\
    CLL\_SUB\_111 & 147.15 - & \textbf{6.94 +} & 807.50 - & 595.82 - & 1080.10 - & 352.00 - & 95.64 \\
    DLBCL & 3.70 + & \textbf{2.68 +} & 296.55 - & 118.89 - & 257.69 - & 34.80 + & 92.32 \\
    GLI\_85 & 47.90 + & \textbf{9.56 +} & 713.75 - & 778.93 - & 1179.69 - & 259.70 + & 330.46 \\
    Leukemia1 & 4.15 + & \textbf{2.35 +} & 246.45 - & 239.63 - & 333.23 - & 102.85 - & 81.87 \\
    Leukemia2 & 4.55 + & \textbf{3.83 +} & 479.25 - & 382.58 - & 770.14 - & 126.65 + & 159.64 \\
    Lung\_Cancer & 42.35 + & \textbf{8.40 +} & 617.15 - & 646.32 - & 1584.38 - & 339.55 - & 157.01 \\
    Lymphoma & 2.00 + & \textbf{1.64 +} & 10.30 + & 205.33 - & 269.03 - & 26.10 + & 39.30 \\
    Nci   & 53.55 + & \textbf{4.10 +} & 336.15 - & 235.19 - & 597.05 - & 232.00 - & 83.83 \\
    Nci9  & 82.05 + & \textbf{6.37 +} & 2992.45 - & 444.54 - & 714.44 - & 1770.90 - & 131.17 \\
    ORL10P & 6.70 + & \textbf{2.53 +} & 172.50 + & 677.44 - & 1475.57 - & 128.95 + & 262.78 \\
    PIE10P & 10.65 + & \textbf{4.13 +} & 429.75 - & 101.47 - & 317.27 - & 41.05 + & 69.03 \\
    PIX10P & 2.50 + & \textbf{1.90 +} & 836.40 - & 766.71 - & 1270.15 - & 85.05 + & 212.77 \\
    Prostate\_GE & 8.70 + & \textbf{3.01 +} & 417.80 - & 175.47 - & 398.11 - & 253.05 = & 119.42 \\
    Prostate\_Tumor & 13.10 + & \textbf{4.98 +} & 414.30 - & 281.16 - & 714.02 - & 109.35 = & 94.77 \\
    Prostate\_6033 & 14.25 + & \textbf{4.08 +} & 232.05 - & 173.69 - & 381.60 - & 118.10 = & 110.77 \\
    SMK\_CAN\_187 & 65.65 = & \textbf{11.21 +} & 876.15 - & 896.52 - & 1391.80 - & 469.45 - & 72.87 \\
    SRBCT & 4.20 + & \textbf{2.59 +} & 334.50 - & 46.77 + & 131.01 - & 57.35 = & 58.98 \\
    TOX\_171 & 45.10 + & \textbf{7.49 +} & 424.50 - & 261.38 - & 515.67 - & 1207.70 - & 166.40 \\
    \hline
    \multicolumn{1}{c}{+/-/=} & \multicolumn{1}{c}{21/2/3} & \multicolumn{1}{c}{26/0/0} & \multicolumn{1}{c}{2/24/0} & \multicolumn{1}{c}{1/25/0} & \multicolumn{1}{c}{0/26/0} & \multicolumn{1}{c}{7/13/6} &  \\
    \hline
    \hline
    \end{tabular}%
    }
  \label{ch5/tab:fr_all}%
\end{table*}%

\subsection{Performance Comparison} 
This subsection compares the proposed MO-FSEMT method with the six comparison algorithms regarding classification performance and computation efficiency. The experimental results of all compared methods are obtained from twenty independent runs. The Wilcoxon rank-sum test~\cite{wilcoxon1970critical} with a significance level of 0.05 is applied to the following results of classification performance, where symbols ``+'', ``-'', and ``='' indicate that the compared method is significantly better than, worse than, and similar to the proposed MO-FSEMT.

\subsubsection{Classification Results}
Classification accuracy and the number of selected features are essential metrics for evaluating the performance of FS methods. Higher classification accuracy and less number of features indicate the superiority of the current algorithm in solving FS problems. In addition, it is worth noting that the number of solutions obtained by the comparison algorithms in each run is not the same, which influences the performance comparison. Specifically, most single-objective algorithms only retain the best FS solution in one run, such as HFS-C-P and MFCSO. Regarding PS-NSGA, although it considers multiple objectives during environmental selection, only the best solution (the individual with the smallest distance among the most accurate individuals) is selected for output after the evolution is completed. In contrast, as single-objective algorithms, PSO-EMT and MTPSO can output more than a single solution since they utilize multitask optimization and retain every best solution for each task. Furthermore, the remaining two multiobjective algorithms, i.e., SM-MOEA and the proposed MO-FSEMT, can obtain a set of non-dominated solutions at a time. To ensure fairness, in each run, the average value and the best value of the classification accuracy of the solution sets obtained by the multi-output algorithms are taken for subsequent comparisons, respectively.

Table~\ref{ch5/tab:meanAcc_all} presents the average classification accuracies (with standard deviation) of all comparative algorithms on 26 datasets based on 20 independent runs. The highest classification accuracy obtained for each dataset is highlighted in bold. As can be seen from Table~\ref{ch5/tab:meanAcc_all}, MO-FSEMT achieves the best results on 18 out of 26 datasets. In the 156 comparisons regarding the significance tests, MO-FSEMT wins 134 wins, draws 14, and losses 8, confirming the effectiveness of the proposed method.
To be specific, the superiority of MO-FSEMT over multiobjective algorithms such as PS-NSGA and SM-MOEA confirms it as an efficient multiobjective algorithm for solving high-dimensional FS problems. On the other hand, the advantages of MO-FSEMT over HFS-C-P, PSO-EMT, MTPSO, and MFCSO validate the effectiveness of the designed EMT framework.
Table~\ref{ch5/tab:maxAcc_all} compares the best classification accuracy values obtained by MO-FSEMT and other comparative algorithms, further verifying its competitiveness. For algorithms capable of outputting multiple solutions, the best classification accuracy results can reflect the exploration capability of the algorithm. According to the statistical data in Table~\ref{ch5/tab:maxAcc_all}, it is evident that MO-FSEMT achieves the best performance with overwhelming superiority, confirming that MO-FSEMT can not only obtain superior trade-off solution sets but also search for even better solutions.

In terms of the average number of features selected by MO-FSEMT and the other six compared algorithms listed in Table~\ref{ch5/tab:fr_all}, it can be observed that PS-NSGA and SM-MOEA obtain smaller feature subsets than MO-FSEMT on most datasets. The main reason is that each objective value of most multiobjective FS methods is equally important during environmental selection and evolution. Furthermore, a specific operator is designed to reduce the number of selected features in SM-MOEA. Although PS-NSGA has a preference for the objective of classification accuracy in selection and ranking, the mutation operator used previously to generate offspring is designed for rapid dimensionality reduction. However, the lower-dimensional feature sets provided by these methods lead to higher classification error rates, as shown in Tables~\ref{ch5/tab:meanAcc_all} and~\ref{ch5/tab:maxAcc_all}.
On the contrary, the single-objective EMT-based FS methods such as PSO-EMT, MTPSO, and MFCSO are underperformed in reducing the number of features, mainly due to their emphasis on weighted objectives prioritizing classification accuracy over feature size. Moreover, in HFS-C-P, only classification accuracy is considered as a single objective for the population, and a three-stage hybrid FS algorithm is designed to decrease the search space, resulting in larger feature sizes.

The comprehensive empirical results presented above confirm the effectiveness of our proposed MO-FSEMT compared to six EC-based FS methods in addressing FS problems with high dimensions by obtaining a set of superior trade-off solutions with few features. The improved performance of MO-FSEMT can be attributed to the synergistic collaboration of various components in the proposed framework, including redundant feature removal, task construction using different problem formulation methods, different fitness calculation strategies in independent solvers, and task-specific knowledge transfer mechanisms. In the following subsection, the effectiveness of these proposed strategies in MO-FSEMT is demonstrated.

\subsubsection{Computation Time}
\begin{figure}[htbp]
    \centering
    \includegraphics[width=0.48\textwidth]{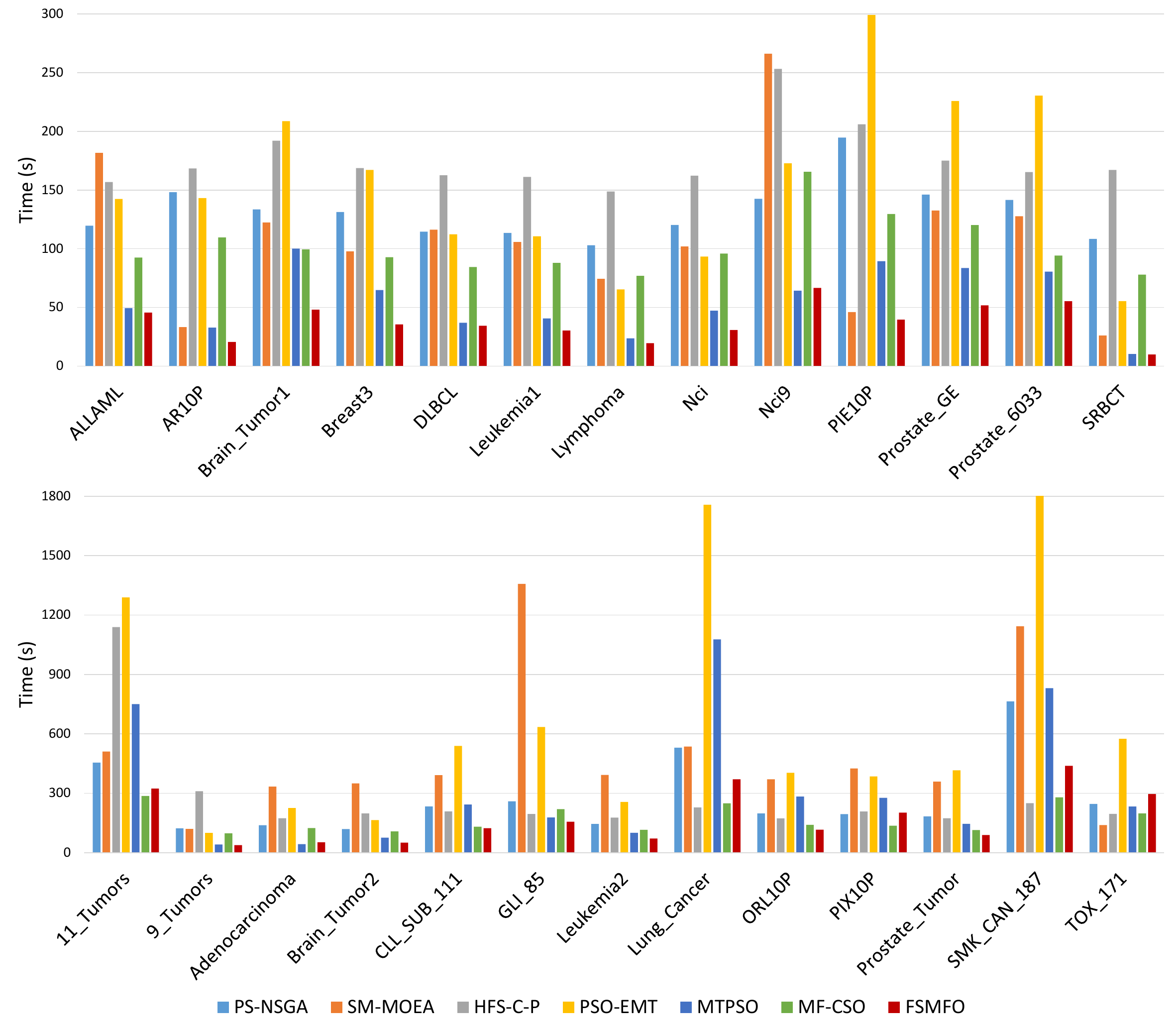}
    \caption{The average computation time of seven compared methods on 26 datasets over 20 independent runs. x-axis: datasets; y-axis: time(s).}
    \label{ch5/fig:time}
\end{figure}

To elucidate the computational complexity of the algorithms, this subsection compares the proposed MO-FSEMT with other EC-based comparison algorithms in terms of runtime under the same settings. Figure~\ref{ch5/fig:time} provides the average CPU running time consumed by seven methods on 26 datasets over 20 independent run times. It can be observed that MO-FSEMT shows the shortest runtime in 19 out of the 26 datasets, indicating that the proposed multi-formulation and multi-solver optimization strategies not only achieve good classification performance but also possess high computational efficiency. Problem formulation can effectively generate multiple promising tasks with fewer features, significantly reducing the computational time required for individual fitness evaluation during the evolutionary process. In addition, optimizing different tasks with multiple solvers based on task characteristics and facilitating knowledge transfer among tasks through the proposed mechanisms can improve the quality of the solution set and enhance search efficiency. Although the computation time of MO-FSEMT is slightly surpassed by other comparative algorithms on 7 out of 26 datasets, MO-FSEMT achieves better classification performance on these datasets. Overall, our proposed MO-FSEMT method strikes the best balance between effectiveness and efficiency among all the algorithms.

\subsection{Ablation Study}
To further analyze the contributions of different components of MO-FSEMT, we conduct ablation experiments to investigate the effects of removing redundant features, different problem formulation methods, different fitness calculation strategies within independent solvers, and task-specific based knowledge transfer mechanisms on improving the performance of solving FS problems. Table~\ref{ch5/tab:meanAcc_ab}, Table~\ref{ch5/tab:maxAcc_ab}, and Table~\ref{ch5/tab:fr_ab} summarize the statistical comparison results of the four groups of ablation experiments in terms of average classification accuracy value, optimal classification accuracy value, and average number of selected features, respectively. 

\subsubsection{The Effect of Redundancy Removal}
To validate the effectiveness of removing irrelevant features in the MO-FSEMT framework, we compare MO-FSEMT and its variant W/O R (without removal) on nine datasets with a significant difference in the number of features after redundancy removal, as listed in Table~\ref{ch5/tab:removalF}.

\linespread{0.9}
\begin{table}[htbp]
  \centering
  \caption{Comparison of the number of full features and the number of related features on nine datasets}
    \begin{tabular}{p{3cm}p{3cm}p{3cm}}
    \hline
    \hline
    \textbf{Dataset} & \multicolumn{1}{c}{\textbf{\#Full of features}} & \multicolumn{1}{c}{\textbf{\#Relevant features}} \\
    \hline
    9\_Tumors & \multicolumn{1}{r}{5726}  & \multicolumn{1}{r}{5714} \\
    Adenocarcinoma & \multicolumn{1}{r}{9672}  & \multicolumn{1}{r}{8795} \\
    ALLAML & \multicolumn{1}{r}{7129}  & \multicolumn{1}{r}{7081} \\
    DLBCL & \multicolumn{1}{r}{5469}  & \multicolumn{1}{r}{5447} \\
    GLI-85  & \multicolumn{1}{r}{22283}  & \multicolumn{1}{r}{22193} \\
    Leukemia2 & \multicolumn{1}{r}{11225}  & \multicolumn{1}{r}{11170} \\
    Nci9  & \multicolumn{1}{r}{9712}  & \multicolumn{1}{r}{8182} \\
    Prostate\_GE  & \multicolumn{1}{r}{5966}  & \multicolumn{1}{r}{5843} \\
    Prostate\_Tumor & \multicolumn{1}{r}{10509}  & \multicolumn{1}{r}{10466} \\
    \hline
    \hline
    \end{tabular}%
  \label{ch5/tab:removalF}%
\end{table}%

The statistical results from the second column of Tables~\ref{ch5/tab:meanAcc_ab}-\ref{ch5/tab:fr_ab} indicate that MO-FSEMT achieved better classification performance and a smaller number of features compared to the variant retained all features on most datasets, particularly on \textit{Adenocarcinoma}, \textit{GLI-85} and \textit{Prostate\_GE} with a large difference in number before and after. This suggests the importance of redundancy removal in improving the performance of MO-FSEMT.

\begin{table*}[htbp]
  \centering
  \caption{Pairwise comparisons between MO-FSEMT and different variants on the average classification accuracy values}
    \scalebox{0.9}{
    \begin{tabular}{l|c|cc|cc|cc|r}
    \hline
    \hline
    \multirow{2}[3]{*}{\textbf{Dataset}} & \textbf{Removal} & \multicolumn{2}{c|}{\textbf{Task Generation }} & \multicolumn{2}{c|}{\textbf{Fitness of Solver}} & \multicolumn{2}{c|}{\textbf{Transfer Strategy}} & \multirow{2}[3]{*}{\textbf{MO-FSEMT}} \\
\cline{2-8}          & \textbf{W/O R} & \textbf{W/O F} & \textbf{W/O C} & \textbf{Fit 1} & \textbf{Fit 2} & \textbf{W/O Tr} & \textbf{Tr\_SBX} & \\
    \hline
    11\_Tumors &    -   & 72.56 - & 81.87 - & 81.81 - & 85.18 = & 85.16 = & 84.58 = & 85.18 \\
    9\_Tumors & 67.13 = & 41.62 - & 61.52 - & 61.88 - & 67.81 = & 71.92 + & 69.85 + & 68.27 \\
    Adenocarcinoma & 64.72 - & 59.40 - & 59.52 - & 64.93 - & 63.58 - & 63.27 - & 66.02 - & 68.68 \\
    ALLAML & 96.22 = & 84.80 - & 95.62 - & 94.97 - & 95.89 - & 97.62 + & 97.90 + & 96.45 \\
    AR10P &    -   & 53.60 - & 78.60 - & 79.14 - & 79.94 - & 80.34 = & 81.10 = & 81.21 \\
    Brain\_Tumor1 &    -   & 74.44 = & 71.62 - & 72.27 - & 75.27 = & 79.06 + & 78.33 + & 74.70 \\
    Brain\_Tumor2 &    -   & 64.33 - & 76.79 - & 80.30 = & 79.99 - & 82.00 = & 82.66 = & 82.28 \\
    Breast3 &    -   & 54.39 - & 59.50 - & 60.17 - & 61.40 - & 68.42 + & 66.96 + & 63.85 \\
    CLL\_SUB\_111 &    -   & 71.64 - & 78.82 - & 80.08 = & 81.21 = & 86.08 + & 84.51 + & 80.51 \\
    DLBCL & 94.34 - & 81.89 - & 94.61 - & 94.14 - & 94.98 = & 95.63 = & 95.39 = & 95.48 \\
    GLI\_85 & 91.41 - & 84.47 - & 89.26 - & 91.87 - & 93.50 = & 96.58 + & 94.75 + & 93.33 \\
    Leukemia1 &    -   & 84.04 - & 93.71 - & 94.18 - & 94.95 - & 95.51 = & 95.48 = & 95.35 \\
    Leukemia2 & 93.90 - & 87.93 - & 94.51 - & 94.50 - & 95.17 - & 95.41 - & 95.82 = & 96.11 \\
    Lung\_Cancer &    -   & 76.16 - & 87.57 - & 88.47 - & 91.04 = & 93.85 + & 92.81 + & 90.77 \\
    Lymphoma &    -   & 99.02 + & 98.79 = & 98.74 = & 98.97 = & 99.55 + & 99.50 + & 98.75 \\
    Nci   &    -   & 67.56 - & 72.66 - & 71.74 - & 75.61 + & 77.34 + & 78.30 + & 74.01 \\
    Nci9  & 77.16 = & 49.96 - & 69.78 - & 69.27 - & 77.68 = & 83.49 + & 82.54 + & 76.64 \\
    ORL10P &    -   & 94.81 - & 98.42 - & 98.89 - & 99.00 - & 99.22 = & 98.90 - & 99.30 \\
    PIE10P &    -   & 99.29 + & 96.41 - & 96.56 - & 98.46 = & 98.75 = & 98.75 = & 98.58 \\
    PIX10P &    -   & 99.00 + & 97.68 - & 98.71 - & 98.77 = & 99.00 + & 99.01 + & 98.95 \\
    Prostate\_GE & 90.79 - & 83.39 - & 63.13 - & 90.12 - & 91.22 - & 92.28 = & 91.54 = & 92.12 \\
    Prostate\_Tumor & 91.06 = & 82.59 - & 89.11 - & 90.35 = & 92.15 + & 93.17 + & 93.22 + & 91.34 \\
    Prostate\_6033 &    -   & 81.85 - & 89.36 - & 90.45 - & 90.97 - & 92.09 = & 91.94 = & 91.80 \\
    SMK\_CAN\_187 &    -   & 63.14 - & 50.43 - & 71.19 - & 72.73 = & 82.54 + & 79.29 + & 73.59 \\
    SRBCT &    -   & 88.33 - & 97.85 - & 97.35 - & 98.19 - & 99.00 + & 98.94 + & 98.58 \\
    TOX\_171 &    -   & 88.01 - & 60.88 - & 84.84 - & 89.55 + & 91.92 + & 92.58 + & 88.60 \\
    \hline
    \multicolumn{1}{c|}{+/-/=} & 0/5/4 & 3/22/1 & 0/25/1 & 0/22/4 & 3/11/12 & 15/2/9 & 15/2/9 & \\
    \hline
    \hline
    \end{tabular}%
    }
  \label{ch5/tab:meanAcc_ab}%
\end{table*}%

\begin{table*}[htbp]
  \centering
  \caption{Pairwise comparisons between MO-FSEMT and different variants on the best classification accuracy values}
    \scalebox{0.9}{
    \begin{tabular}{l|c|cc|cc|cc|r}
    \hline
    \hline
    \multirow{2}[3]{*}{\textbf{Dataset}} & \textbf{Removal} & \multicolumn{2}{c|}{\textbf{Task Generation }} & \multicolumn{2}{c|}{\textbf{Fitness of Solver}} & \multicolumn{2}{c|}{\textbf{Transfer Strategy}} & \multirow{2}[3]{*}{\textbf{MO-FSEMT}} \\
\cline{2-8}          & \textbf{W/O R} & \textbf{W/O F} & \textbf{W/O C} & \textbf{Fit 1} & \textbf{Fit 2} & \textbf{W/O Tr} & \textbf{Tr\_SBX} & \\
    \hline
    11\_Tumors &    -   & 76.68 - & 91.70 - & 92.44 = & 91.66 = & 89.67 - & 89.64 - & 92.70 \\
    9\_Tumors & 82.05 = & 46.69 - & 80.10 = & 79.77 - & 81.37 = & 80.03 = & 81.66 = & 82.01 \\
    Adenocarcinoma & 82.64 = & 66.13 - & 79.69 = & 82.17 = & 79.93 - & 69.67 - & 74.28 - & 84.62 \\
    ALLAML & 100.00 = & 88.09 - & 100.00 = & 100.00 = & 100.00 = & 99.88 = & 100.00 = & 100.00 \\
    AR10P &    -   & 56.15 - & 87.10 = & 87.88 = & 85.88 - & 82.95 - & 84.75 - & 87.80 \\
    Brain\_Tumor1 &    -   & 78.61 - & 85.91 = & 87.02 = & 87.60 = & 84.90 - & 84.65 - & 87.33 \\
    Brain\_Tumor2 &    -   & 68.54 - & 92.44 = & 95.10 + & 89.75 - & 89.21 - & 89.25 - & 93.48 \\
    Breast3 &    -   & 58.36 - & 75.67 = & 75.47 = & 73.16 - & 76.78 = & 77.18 = & 77.07 \\
    CLL\_SUB\_111 &    -   & 75.91 - & 88.92 = & 89.39 = & 89.45 = & 90.24 = & 89.76 = & 89.70 \\
    DLBCL & 100.00 = & 86.03 - & 100.00 = & 100.00 = & 100.00 = & 100.00 = & 100.00 = & 100.00 \\
    GLI\_85 & 98.38 = & 85.73 - & 97.57 - & 98.64 = & 98.81 = & 98.54 = & 98.59 = & 98.71 \\
    Leukemia1 &    -   & 87.44 - & 100.00 = & 99.92 = & 100.00 = & 100.00 = & 99.92 = & 100.00 \\
    Leukemia2 & 100.00 = & 89.31 - & 100.00 = & 100.00 = & 99.92 = & 98.61 - & 98.92 - & 100.00 \\
    Lung\_Cancer &    -   & 79.16 - & 97.09 = & 97.47 = & 96.81 = & 96.19 - & 95.60 - & 97.16 \\
    Lymphoma &    -   & 99.13 - & 100.00 = & 100.00 = & 100.00 = & 100.00 = & 100.00 = & 100.00 \\
    Nci   &    -   & 71.97 - & 86.82 = & 87.03 = & 85.47 = & 84.19 - & 85.40 = & 86.62 \\
    Nci9  & 89.08 = & 56.30 - & 86.24 - & 88.14 = & 89.36 = & 89.23 = & 89.42 = & 89.91 \\
    ORL10P &    -   & 95.55 - & 100.00 = & 100.00 = & 99.95 = & 99.90 = & 100.00 = & 100.00 \\
    PIE10P &    -   & 99.30 - & 99.44 - & 99.62 = & 99.24 - & 99.28 - & 99.27 - & 99.77 \\
    PIX10P &    -   & 99.00 - & 99.70 = & 99.80 = & 99.30 = & 99.05 - & 99.55 = & 99.55 \\
    Prostate\_GE & 96.85 = & 87.02 - & 96.37 = & 96.95 = & 96.57 - & 96.16 - & 95.77 - & 97.27 \\
    Prostate\_Tumor & 96.08 - & 86.18 - & 96.27 = & 96.74 = & 96.91 = & 95.79 - & 96.50 = & 96.57 \\
    Prostate\_6033 &    -   & 85.32 - & 96.19 = & 96.96 = & 96.25 - & 95.87 - & 95.91 - & 97.04 \\
    SMK\_CAN\_187 &    -   & 65.54 - & 81.80 = & 81.07 = & 80.36 = & 86.42 + & 84.07 + & 81.90 \\
    SRBCT &    -   & 90.46 - & 100.00 = & 100.00 = & 100.00 = & 100.00 = & 100.00 = & 100.00 \\
    TOX\_171 &    -   & 90.32 - & 94.40 = & 94.41 = & 93.89 = & 93.99 = & 95.49 + & 94.46 \\
    \hline
    \multicolumn{1}{c|}{+/-/=} & 0/1/8 & 0/26/0 & 0/4/22 & 1/1/24 & 0/7/19 & 1/13/12 & 2/10/14 &  \\
    \hline
    \hline
    \end{tabular}%
    }
  \label{ch5/tab:maxAcc_ab}%
\end{table*}%
\begin{table*}[htbp]
  \centering
  \caption{Pairwise comparisons between MO-FSEMT and different variants on the number of the selected features}
    \scalebox{0.9}{
    \begin{tabular}{l|c|cc|cc|cc|r}
    \hline
    \hline
    \multirow{2}[3]{*}{\textbf{Dataset}} & \textbf{Removal} & \multicolumn{2}{c|}{\textbf{Task Generation }} & \multicolumn{2}{c|}{\textbf{Fitness of Solver}} & \multicolumn{2}{c|}{\textbf{Transfer Strategy}} & \multirow{2}[3]{*}{\textbf{MO-FSEMT}} \\
\cline{2-8}          & \textbf{W/O R} & \textbf{W/O F} & \textbf{W/O C} & \textbf{Fit 1} & \textbf{Fit 2} & \textbf{W/O Tr} & \textbf{Tr\_SBX} & \\
    \hline
    11\_Tumors &    -   & 4936.99 - & 189.25 + & 185.51 + & 289.42 - & 404.44 - & 367.83 - & 271.78 \\
    9\_Tumors & 115.07 = & 2186.87 - & 84.90 + & 83.22 + & 124.47 - & 174.39 - & 153.22 - & 113.48 \\
    Adenocarcinoma & 26.86 = & 3445.63 - & 10.55 + & 19.17 = & 24.42 = & 423.93 - & 375.49 - & 21.59 \\
    ALLAML & 66.11 - & 2751.39 - & 54.80 + & 41.61 + & 64.19 = & 197.00 - & 168.97 - & 60.61 \\
    AR10P &    -   & 841.88 - & 38.07 + & 44.44 + & 71.19 = & 149.58 - & 145.43 - & 60.90 \\
    Brain\_Tumor1 &    -   & 2246.45 - & 34.19 + & 31.47 + & 51.12 = & 210.26 - & 158.72 - & 50.17 \\
    Brain\_Tumor2 &    -   & 4022.73 - & 78.14 + & 73.11 + & 188.65 - & 508.90 - & 421.94 - & 111.61 \\
    Breast3 &    -   & 1817.78 - & 31.47 + & 23.95 + & 47.47 = & 183.67 - & 134.28 - & 46.64 \\
    CLL\_SUB\_111 &    -   & 4381.12 - & 53.75 + & 55.33 + & 83.19 = & 412.44 - & 367.14 - & 95.64 \\
    DLBCL & 91.31 = & 2112.69 - & 70.99 + & 61.83 + & 94.96 = & 211.53 - & 199.27 - & 92.32 \\
    GLI\_85 & 312.84 = & 8839.53 - & 158.43 + & 219.33 + & 300.18 = & 629.25 - & 561.29 - & 330.46 \\
    Leukemia1 &    -   & 2024.69 - & 59.91 + & 53.58 + & 84.06 = & 150.98 - & 136.15 - & 81.87 \\
    Leukemia2 & 156.34 = & 4314.71 - & 116.20 + & 110.03 + & 168.51 = & 329.93 - & 297.16 - & 159.64 \\
    Lung\_Cancer &    -   & 4799.65 - & 97.00 + & 90.41 + & 150.22 = & 597.48 - & 492.87 - & 157.01 \\
    Lymphoma &    -   & 1590.41 - & 34.98 + & 34.09 + & 47.16 - & 222.25 - & 211.84 - & 39.30 \\
    Nci   &    -   & 1966.48 - & 60.19 + & 53.71 + & 91.16 - & 131.36 - & 113.33 - & 83.83 \\
    Nci9  & 147.13 - & 3426.21 - & 83.29 + & 84.53 + & 140.90 = & 248.40 - & 207.28 - & 131.17 \\
    ORL10P &    -   & 3950.13 - & 164.58 + & 187.23 + & 270.24 = & 562.47 - & 485.63 - & 262.78 \\
    PIE10P &    -   & 868.69 - & 39.96 + & 49.85 + & 87.44 - & 114.05 - & 97.66 - & 69.03 \\
    PIX10P &    -   & 3955.71 - & 105.51 + & 191.53 = & 293.12 - & 2889.11 - & 2278.79 - & 212.77 \\
    Prostate\_GE & 132.15 - & 2270.88 - & 62.70 + & 69.80 + & 124.70 = & 181.48 - & 169.61 - & 119.42 \\
    Prostate\_Tumor & 99.26 = & 4113.11 - & 50.54 + & 57.23 + & 104.83 = & 284.52 - & 245.16 - & 94.77 \\
    Prostate\_6033 &    -   & 2322.60 - & 50.58 + & 62.10 + & 101.97 + & 159.24 - & 145.91 - & 110.77 \\
    SMK\_CAN\_187 &    -   & 7767.83 - & 62.81 = & 42.44 + & 84.35 - & 342.26 - & 281.26 - & 72.87 \\
    SRBCT &    -   & 807.22 - & 48.84 + & 40.67 + & 62.64 - & 70.12 - & 63.09 - & 58.98 \\
    TOX\_171 &    -   & 2146.66 - & 136.15 = & 120.49 + & 184.57 = & 545.19 - & 451.86 - & 166.40 \\
    \hline
    \multicolumn{1}{c|}{+/-/=} & 0/3/6 & 0/26/0 & 24/0/2 & 24/0/2 & 1/9/16 & 0/26/0 & 0/26/0 &  \\
    \hline
    \hline
    \end{tabular}%
    }
  \label{ch5/tab:fr_ab}%
\end{table*}%

\subsubsection{The Effect of Problem Formulation}
To investigate the contributions of different problem formulations in MO-FSEMT to improving classification accuracy and reducing the number of selected features, we designed two variants of MO-FSEMT that employ a single approach for problem formulation. W/O F (without filtering) represents the variant that only utilizes the clustering method for generating auxiliary tasks, while W/O C (without clustering) represents the one that only utilizes filtering. The columns 3-4 in Tables~\ref{ch5/tab:meanAcc_ab}-\ref{ch5/tab:fr_ab} present the statistical results of their performance comparison.

It can be observed that except for a slight improvement in the average classification accuracy value on a few datasets, both W/O F and W/O C exhibit a significant decrease in classification accuracy compared to MO-FSEMT. This indicates both formulation approaches are effective in improving classification accuracy. On the other hand, the results of the feature size of W/O F are significantly higher than in the proposed MO-FSEMT across all datasets, while W/O C shows slightly better results. This suggests that the filtering-based formulation approach is more effective in reducing the number of selected features.

By leveraging the advantages of different problem formulations, the proposed MO-FSEMT achieves the overall optimal classification accuracy and feature selection performance.

\subsubsection{The Effect of Independent Solver}
To evaluate the effectiveness of the proposed multi-solver-based multitask optimization in MO-FSEMT, we design two variants of MO-FSEMT, namely Fit1 and Fit2, by adjusting different fitness calculation strategies in the independent solvers. The Fit1 variant considers both the classification error rate and the feature ratio when calculating the fitness in independent solvers of the original task and the auxiliary tasks, while the Fit2 variant only considers the error rate objective value on all tasks.

As shown in columns 5-6 of Tables~\ref{ch5/tab:meanAcc_ab}-\ref{ch5/tab:fr_ab}, the Fit1 variant, which considers the feature ratio in the fitness calculation of auxiliary tasks with smaller feature size, selects fewer features than MO-FSEMT on most datasets. However, due to the dominance of this objective, the average classification accuracy results of Fit1 decrease compared to MO-FSEMT. Since the Fit2 variant only considers the error rate, higher classification accuracy can be achieved on some datasets, but slightly more selected features could be possessed.

Overall, the proposed MO-FSEMT, which considers reducing the classification error and feature ratio simultaneously in the original task while focusing more on classification error in the auxiliary tasks with smaller feature size, achieves a more balanced classification performance.

\subsubsection{The Effect of Knowledge Transfer Mechanism}
To investigate the impact of the proposed task-specific knowledge transfer mechanism, we introduce two variants of MO-FSEMT. W/O Tr (without transfer) represents the variant without knowledge transfer across related tasks, while Tr\_SBX adopts the traditional chromosome crossover and mutation approach to transfer information. The comparative results are presented in columns 7-8 of Tables~\ref{ch5/tab:meanAcc_ab}-\ref{ch5/tab:fr_ab}.

Both W/O Tr and Tr\_SBX variants show similar behavior. Slightly higher average classification accuracy results are achieved by them than MO-FSEMT on half of the datasets, but conversely, they perform worse in terms of the best classification accuracy. Considering the number of selected features, both variants resulted in a significantly larger feature subset than MO-FSEMT across all datasets, with W/O Tr selecting more features than Tr\_SBX.
These results indicate that allowing each task to evolve independently without sharing useful knowledge leads to stagnation in local optima, obtaining solutions that only meet part of the objectives rather than better and balanced solutions. However, employing the traditional crossover and mutation method for knowledge transfer neglects the specific characteristics of different tasks, providing limited improvements to address the drawbacks above.

The proposed task-specific knowledge transfer mechanism effectively leverages the distinct information provided by different tasks, enabling better utilization of the advantages of each task. For FS problems with multiple conflicting objectives, MO-FSEMT demonstrates superior trade-off performance overall.

\subsection{Case Study on Higher-Dimensional Data of Glioma Samples}

In this section, we conducted a case study on higher-dimensional expression data (i.e., involving 49151 features) of glioma samples to further validate the effectiveness of the MO-FSEMT framework in real-world FS problems.
This dataset is used to analyze the effect of glioma-derived stem cell factors in brain tumor angiogenesis~\cite{sun2006neuronal}.

Table~\ref{ch5/tab:high} provides a performance comparison of MO-FSEMT and five available FS methods on the higher-dimensional case study. To ensure fairness, for algorithms that generate multiple solutions, we separately compute the average value (MeanAcc) and the best value (MaxAcc) of classification accuracy for comparison. It can be observed that MO-FSEMT achieves the highest classification accuracy while maintaining a smaller feature size, thus validating the ability of the proposed framework to handle higher-dimensional FS problems.

\begin{table}[htbp]
  \centering
  \caption{Comparison between MO-FSEMT and five available FS methods on the higher-dimensional dataset}
  \scalebox{0.89}{
    \begin{tabular}{p{1.2cm}p{1.5cm}p{1.54cm}p{1.54cm}p{1.4cm}}
    \hline
    \hline
    \textbf{Dataset} & \textbf{Algorithms} & \textbf{MeanAcc} & \textbf{MaxAcc} & \textbf{FeatureSize} \\
    \hline
    \multicolumn{1}{l}{\multirow{6}[2]{*}{Case\_Study}} & PS-NSGA & 63.61 (2.90) - & 63.61 (2.90) - & \textbf{310.45 +} \\
          & HFS-C-P & 61.73 (2.77) - & 61.73 (2.77) - & 3478.25 - \\
          & PSO-EMT & 62.56 (2.73) - & 67.17 (2.90) - & 10062.49 - \\
          & MTPSO & 63.17 (2.31) - & 72.47 (2.55) - & 4619.66 - \\
          & MF-CSO & 63.25 (2.23) - & 63.25 (2.23) - & 1011.00 - \\
          & MO-FSEMT & \textbf{70.03 (1.46)} & \textbf{78.03 (1.93)} & 565.69 \\
    \hline
    \hline
    \end{tabular}%
    }
  \label{ch5/tab:high}%
\end{table}%

\section{Conclusion and Future Work}
In this paper, we have proposed the MO-FSEMT framework to address the multiobjective high-dimensional feature selection problems. This framework designs a multitask evolutionary search paradigm, which integrates filtering-based and clustering-based methods to reformulate the original high-dimensional FS problem into simplified auxiliary tasks. By incorporating multiple evolutionary solvers with problem-specific biases and search preferences and enabling knowledge transfer across tasks, better search performance and high-quality solutions can be achieved. The proposed task-specific-based knowledge transfer mechanism considers the task-specific information of individuals to be transferred to enhance the effectiveness of knowledge transfer. Comprehensive experimental results on 26 real high-dimensional datasets demonstrate that MO-FSEMT outperforms other state-of-the-art EC-based FS methods regarding classification accuracy, selected feature quantity, and efficiency. Moreover, extensive ablation experiments further confirm the contribution of key components in the proposed framework to improving classification performance.

In the future, we would like to further explore performance enhancements for the proposed MO-FSEMT framework, including designing alternative problem formulations, investigating the performance of solvers with different search biases, and exploring other transferable knowledge. Notably, the current framework primarily generates auxiliary tasks by simplifying the search space. Future research could construct problem formulations by altering the objective function to capture diverse aspects of feature relevance in high-dimensional datasets, such as dependency, redundancy, and discriminability. Another potential direction for improvement could be investigating ways to handle noise and incomplete data within the MO-FSEMT framework, aiming to improve the reliability and robustness of feature selection in high-dimensional settings. Additionally, the framework could be extended to handle various data types, such as text or image data, to adapt to the demands of real-world applications.

\begin{bibliographystyle}{IEEEtran}

\begin{bibliography}{IEEEabrv,myref}

\end{bibliography}

\end{bibliographystyle}

\end{document}